\documentclass[10pt,twocolumn,letterpaper]{article}

\usepackage[pagenumbers]{cvpr}      % To produce the REVIEW version
\usepackage[dvipsnames,table]{xcolor}
\usepackage{graphicx}
\usepackage{amsmath}
\usepackage{amssymb}
\usepackage{booktabs}
\usepackage{lipsum}
\usepackage{float}
\usepackage{array}
\usepackage{booktabs}
\usepackage{times}
\usepackage{epsfig}
\usepackage{wrapfig}
\usepackage{amsmath,amssymb,amsthm}
\usepackage{algorithm,algorithmicx,algpseudocode}
\usepackage{bm,xspace}
\usepackage{comment}
\usepackage{multirow}
\usepackage{balance}
\usepackage{url}
\usepackage{booktabs}
\usepackage{etoolbox,siunitx}
\usepackage{calc}
\usepackage{pifont,hologo}
\usepackage{color}
\usepackage{adjustbox}
\usepackage[normalem]{ulem}  % \xout

\newcommand{\tablestyle}[2]{\setlength{\tabcolsep}{#1}\renewcommand{\arraystretch}{#2}\centering\footnotesize}

\definecolor{cvprblue}{rgb}{0.21,0.49,0.74}
\usepackage[pagebackref,breaklinks,colorlinks,citecolor=cvprblue]{hyperref}

\usepackage{textpos}
\usepackage{makecell}

\newcommand{\faCry}{\includegraphics[height=2ex]{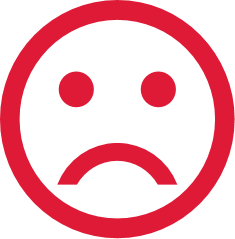}}
\newcommand{\faSmile}{\includegraphics[height=2ex]{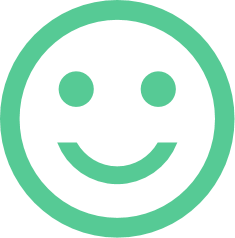}}
\usepackage{pifont}
\newcommand{\methodname}{Alpha-CLIP\xspace}

\newcommand\blfootnote[1]{%
  \begingroup
  \renewcommand\thefootnote{}\footnote{#1}%
  \addtocounter{footnote}{-1}%
  \endgroup
}

\def\paperID{896} % *** Enter the Paper ID here
\def\confName{CVPR}
\def\confYear{2024}

\title{
\raisebox{-1.1ex}{\protect\includegraphics[height=2.7\fontcharht\font`\B]{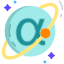}}
Alpha-CLIP: A CLIP Model Focusing on Wherever You Want}

\author{
Zeyi Sun$^{*1,4}$, Ye Fang$^{*2,4}$, Tong Wu$^{3}$, Pan Zhang$^{4}$, Yuhang Zang$^{4}$, \\ Shu Kong$^{5}$, Yuanjun Xiong$^{6}$, Dahua Lin$^{3,4}$, Jiaqi Wang$^{\dagger4}$\\
$^1$Shanghai Jiao Tong University \quad 
$^2$Fudan University \quad
$^3$The Chinese University of Hong Kong \quad  \\
$^4$Shanghai AI Laboratory \quad $^5$University of Macau \quad $^6$MThreads, Inc.\\
{\tt\small szy2023@sjtu.edu.cn, \{fangye, zhangpan, zangyuhang, wangjiaqi\}@pjlab.org.cn}\\
{{\tt\small \url{https://aleafy.github.io/alpha-clip}}
\vspace{-10mm}
}
}
\begin{document}

\twocolumn[{%
\renewcommand\twocolumn[1][]{#1}%
\maketitle
\vspace{-2mm}
\begin{center}
\centering
\includegraphics[width=1.0\textwidth]{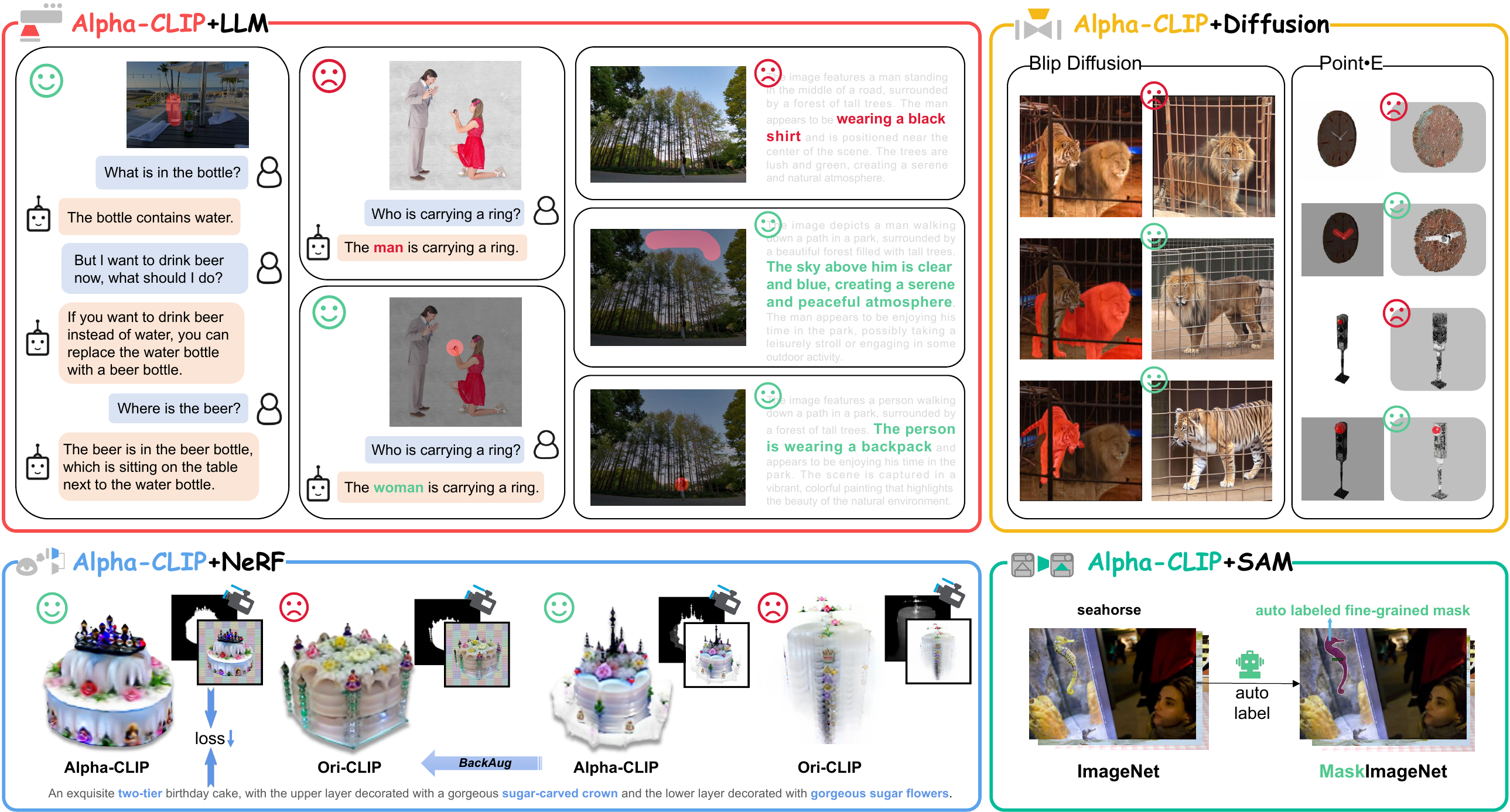}
\vspace{-6mm}
\captionof{figure}{\textbf{Usage of our proposed Alpha-CLIP}.
% Prompting CLIP with the region to be focused through alpha channel, Alpha-CLIP can replace original CLIP in a wide range of task like region focused diffusion, region focused captioning\shu{``bottle'' is mis-spelled!}, 3D object generation and region focused recognition.
% \shu{Use tricks to note failure cases when using CLIP. Current figure doesn't tell readers which are failure cases and which are produced by a typical CLIP.
% }
 Our Alpha-CLIP can seamlessly replace the original CLIP in a wide range of tasks to allow the whole system to focus on any specified region given by points, strokes or masks. Cases marked with \faCry\hspace{2px}are generated with the original CLIP. Cases marked with \faSmile\hspace{2px}are generated with our Alpha-CLIP. All cases shown here are made simply by replacing the original CLIP of the system with a plug-in Alpha-CLIP without further tuning.
% \shu{This figure only showcases what Alpha-CLIP can be used but does not talk about (1) motivations, (2) challenges, and/or (3) technical contributions. For example, you can talk about the motivation --- why developing Alpha-CLIP; challenge in developing Alpha-CLIP, e.g., data is the issue; technical contributions --- obtaining data using existing models and publicly available data. You can highlight its applications in diverse downstream tasks lastly. You can also use A-D as indices to subfigures and refer to them when mentioning the tasks.}
}
\label{fig:teaser}
\vspace{-3pt}
\end{center}%
% \begin{center}
% \centering
% \includegraphics [width=0.95\textwidth]{figure/fig1.pdf}
% \captionof{figure}{\textbf{Qualitative results of our SAM3D.} The first subfigure demonstrate the 3D scene input. The second subfigure is the segmentation masks predicted by SAM3D. The third subfigure is refined masks generated by ensembling the SAM result and the over-segmentation result. The last subfigure is the ground-truth segmentation labels in ScanNet~\cite{dai2017scannet}. }
% \label{fig:fig1}
% \end{center}%
}]
\maketitle
\blfootnote{$^*$ Equal contribution.\ \ $\dagger$ Corresponding authors.}
\begin{abstract}

Contrastive Language-Image Pre-training (CLIP) plays an essential role in extracting valuable content information from images across diverse tasks. It aligns textual and visual modalities to comprehend the entire image, including all the details, even those irrelevant to specific tasks.
However, for a finer understanding and controlled editing of images, it becomes crucial to focus on specific regions of interest, which can be indicated as points, masks, or boxes by humans or perception models.
To fulfill the requirements, we introduce Alpha-CLIP, an enhanced version of CLIP with an auxiliary alpha channel to suggest attentive regions and fine-tuned with constructed millions of RGBA region-text pairs. Alpha-CLIP not only preserves the visual recognition ability of CLIP but also enables precise control over the emphasis of image contents. It demonstrates effectiveness in various tasks, including but not limited to open-world recognition, multimodal large language models, and conditional 2D / 3D generation. It has a strong potential to serve as a versatile tool for image-related tasks. 

\end{abstract}    
% \begin{figure*}[!t]
%     \centering
%     \includegraphics[width=\linewidth]{pointblip/figure/arch1.pdf}
%     \caption{\textbf{ Two-Stage pre-training model architecture and objectives of PointBLIP. } We utilize universal encoders to encode 3D point clouds, texts and queries, while our training objectives differ during different stages. In the initial stage, we collaboratively optimize three objectives that guide the queries in extracting 3D vision representations most relevant to the text. Subsequently, in the second stage, we employ a decoder-based LLM to conduct Language Modeling tasks. }
%     \label{fig:arch}
% \end{figure*}
{
\setlength{\tabcolsep}{4mm}
\begin{table*}[t]
  \centering
  \tablestyle{6pt}{1.1}
  \vspace{-2mm}
  \scalebox{0.9}{
    \begin{tabular}{lllll} 
    \toprule
    \textbf{Domains} & \textbf{Components} &  \textbf{Tasks} & \textbf{Methods} & \textbf{Advantages over the original CLIP} \\
    \cmidrule{1-5}
    \multirow{2}[0]{*}{Image Recognition} & Alpha-CLIP & \makecell[l]{Zero-shot Classification \\ Zero-shot REC }  & - & \makecell[l]{Superior classification accuracy \\ Excellent region-text comprehension ability} \\
    \cmidrule{2-5}
    & Alpha-CLIP + SAM & Data Engine for OVD & Detic~\cite{Detic}  & Higher OVD mAP \\
    \cmidrule{1-5}
     MLLM & Alpha-CLIP + LLM  & VQA, Captioning & BLIP-2~\cite{BLIP-2}, LLaVA-1.5~\cite{llava-1.5}  & \makecell[l]{Region-focused captioning / VQA \\ Eliminating hallucinations \\ Reducing model bias} \\
    \cmidrule{1-5}
    2D Generation & Alpha-CLIP + Diffusion & Image Variation & BLIP-Diffusion~\cite{BLIP-Diffusion}  & \makecell[l]{Controllable generation \\ Enabling subject-driven generation in complex images} \\
    \cmidrule{1-5}
    \multirow{2}[0]{*}{3D Generation} & Alpha-CLIP + Diffusion & Generalized Image-to-3D & Point-E~\cite{Point-E}  & Rectifying absent parts \\
    \cmidrule{2-5}
     & Alpha-CLIP + NeRF & Optimized Image-to-3D & PureCLIPNeRF~\cite{PureCLIPNeRF}  &  Improved 3D optimization results\\
    \bottomrule
    \end{tabular}%
    }
    \vspace{-3mm}
    \caption{\textbf{Downstream tasks of Alpha-CLIP and their advantages over the original CLIP}}
  \label{tab:tasks}%
  \vspace{-5mm}
\end{table*}%
}

\section{Introduction}
\label{sec:introduction}

Recent advances in Contrastive Language-Image Pre-training (CLIP) ~\cite{openclip,CLIP} and its diverse variants ~\cite{EVA-CLIP,Dong_2023_CVPR,Li_2023_CVPR} have established a robust framework for extracting semantically coherent features from both images and text. These features aim to capture all the semantic details within images, exhibiting potent representation capabilities and exceptional generalizability, making them versatile in a variety of downstream tasks, such as open-world recognition~\cite{ViLD,ODISE,maskQCLIP,SAN, OvarNet}, Multimodal Large Language Models (MLLMs)~\cite{BLIP-2,kosmos-1,Kosmos-2,LLaVA, otter, emu, qwen, llava-1.5, zhang2023internlm}, and 2D / 3D generation~\cite{BLIP-Diffusion,ip-adapter,dalle2,dream_fields,CLIP-Mesh,PureCLIPNeRF,Point-E}.

% CLIP has the ability to represent all semantic contents within images, owing to its training in contrasting learning between the image and full image descriptions. Although CLIP captures all the details, not all of them are useful for specific tasks. Certain tasks may require a focus on regions of interest, which can be indicated by humans or perception models such as SAM~\cite{SAM} and proposal networks~\cite{xxx}. These region hints can be used for open-world recognition~\cite{xxx}, captioning~\cite{xxx}, and VQA~\cite{xxx} to enhance image understanding. Alternatively, it is possible to generate content that focuses on a specific region ~\cite{xxx} or refine details ~\cite{xxx} by providing attentive clues.

While CLIP captures the content of the entire image, it is also crucial to focus on the regions of interest to enable a finer understanding~\cite{hossain2019comprehensive,kumar2019region,qiao2020referring, su2023referring,zhou2023openannotate3d, kender2023g2l} and controllable content generation~\cite{wei2023elite, shi2023instantbooth,CLIP-Mesh,PureCLIPNeRF}. These regions can be specified by points, masks, or boxes via human interaction or perception models (e.g.,  SAM~\cite{SAM}, GLIP~\cite{GLIP} and proposal networks~\cite{mattnet}).

\begin{figure*}[t]
    \centering
    \includegraphics[width=\linewidth]{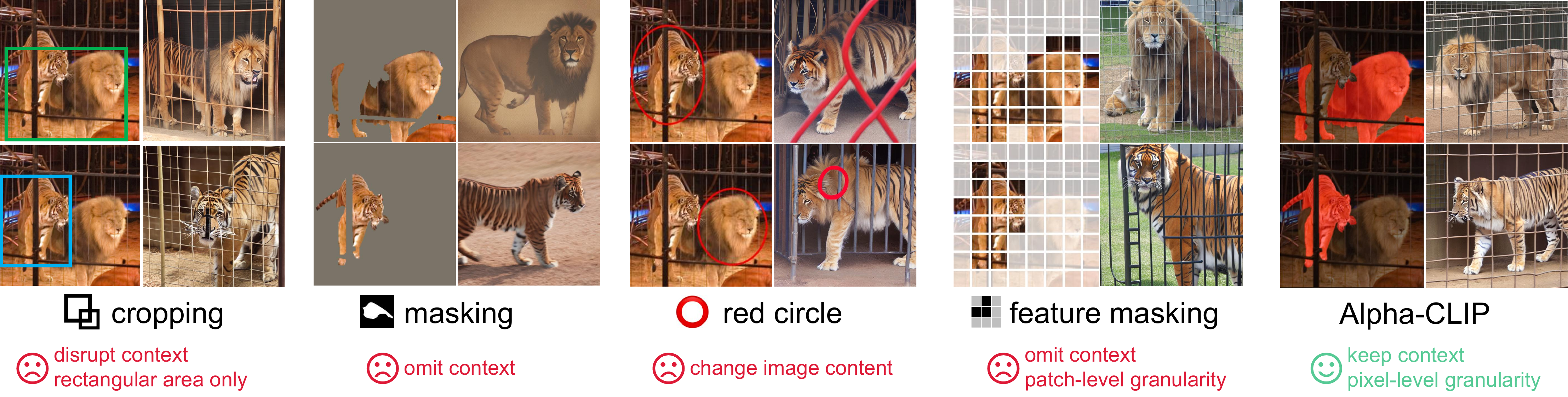}
    \vspace{-7mm}
    \caption{
    % \textbf{Alpha-CLIP compared with other simple methods of region focusing in BLIP-Diffusion~\cite{BLIP-Diffusion}}. The fine-grained region focus ability of Alpha-CLIP produces better results than  other methods that adopts original CLIP.
    \textbf{Alpha-CLIP vs. other  methods of region-focusing} for image generation using BLIP-Diffusion~\cite{BLIP-Diffusion}. 
    The fine-grained region focusing ability of Alpha-CLIP produces better results than these methods that adopt the original CLIP.
    % The lower half shows more results of Alpha-CLIP. 
    % The first row per three is the original BLIP-Diffusion generated images. Other rows represent the outcomes of Alpha-CLIP with highlighted region marked in red.
    }
    \label{fig:region_image_generation}
    \vspace{-5mm}
\end{figure*}

To fulfill the demands of downstream tasks, researchers have attempted to acquire region-focused CLIP features using two primary strategies. The first method is to exclude non-relevant areas by cropping the regions of interest into distinct patches~\cite{zhao2022exploiting,ReCLIP,OvarNet,zhong2022regionclip} or applying masking to the irrelevant parts of images~\cite{MaskAdaptedCLIP}, features~\cite{wei2023elite,MaskAdaptedCLIP}, and attention masks~\cite{MaskCLIP,maskQCLIP}. However, this approach disrupts (in cropping) and omits (in masking) contextual information, which is crucial for precise image understanding and reasoning. The second method is to highlight the regions of interest by circles~\cite{circleCLIP} or mask contour~\cite{yang2023fine} on the images fed to CLIP. Although user-friendly, it changes the original content of the images, which will result in undesirable recognition and generation results (cf. ~\cref{fig:region_image_generation}).
% Empirical comparisons illustrating the outcomes of these methods are available in Figure~\ref{fig:region_image_generation}.

% ~\tong{Maybe we'll need a figure somewhere in the main paper to support this claim, to directly compare our alpha map with the red circle by some examples.}
% subsequently leading to compromised performance.

%Different from existing works, we take a step further to attempt to incorporate the regions of interest as prompts into the CLIP~\cite{CLIP}, enabling it to focus more on region-specific semantic information. We demonstrate that by adding an additional alpha channel along with the RGB channels, we can effectively convey this focusing information to  CLIP~\cite{CLIP}.

To achieve region focus without hurting original image,  we propose Alpha-CLIP, which improves CLIP~\cite{CLIP} by incorporating regions of interest through an additional alpha channel input. Along with the RGB channels, the introduced alpha channel enables the Alpha-CLIP to focus on designated areas while maintaining an awareness of the contextual information. While initialized with the CLIP ~\cite{CLIP} model, the training of Alpha-CLIP still requires a large set of region-text paired data. 
By harnessing the Segment Anything Model (SAM)~\cite{SAM} and multimodal large models for image captioning, such as BLIP-2~\cite{BLIP-2}, we develop an effective pipeline to generate millions of region-text pairs that are readily convertible to RGBA-text data. After training with a mixture of region-text pairs and image-text pairs, Alpha-CLIP can focus on the specific regions while maintaining the visual recognition accuracy of CLIP~\cite{CLIP}.

%Furthermore, in the vast downstream tasks of CLIP~\cite{CLIP}, we select several representative scenarios to illustrate the effectiveness of Alpha-CLIP. 

Alpha-CLIP can enhance CLIP across a wide array of downstream tasks, applying a plug-and-play methodology that permeates diverse domains, spanning from perception to generation in 2D and 3D applications, as shown in ~\cref{fig:teaser} and ~\cref{tab:tasks}.
%including data engine, as shown in Figure.~\ref{fig:teaser} and Table~\ref{tab:tasks}. 
Specifically, \textbf{1) Image Recognition}: Alpha-CLIP not only maintains the visual recognition ability of the original CLIP but also boosts the capability of region-based recognition. Specifically, when provided with ground-truth region to focus on, Alpha-CLIP achieves 4.1\% improvement in top-1 accuracy on zero-shot ImageNet classification task. This superior region-based recognition ability helps downstream tasks like Referring Expression Comprehension(REC)~\cite{ReCLIP} or serves as data engine for Open Vocabulary Detection(OVD)~\cite{Detic}. \textbf{2) Serving as vision backbone for MLLM}: In conjunction with a large language model, Alpha-CLIP becomes capable of facilitating region level captioning and VQA within a MLLM framework. This integration significantly mitigates the occurrences of hallucinations (\eg, black shirt) and diminishes model bias (\eg, man carrying a ring). \textbf{3) 2D generation}: When integrated with a diffusion model, Alpha-CLIP enhances the controllability of BLIP-Diffusion~\cite{BLIP-Diffusion} in image variation tasks. In addition, it enables the extraction of subjects from complex images for subject-driven generation, surmounting an obstacle encountered when deploying BLIP-Diffusion with the original CLIP, which only supports single subjects in simplistic images. \textbf{4) 3D generation}: In addition to the capabilities in 2D generation, Alpha-CLIP exhibits proficiency in 3D generation as well. It can be effectively deployed in conjunction with a diffusion model, such as Point-E~\cite{Point-E}, to enhance the quality of 3D object generation. Additionally, it can be utilized with NeRF~\cite{NeRF}, exemplified by PureCLIPNeRF~\cite{PureCLIPNeRF}, to optimize the creation of superior 3D objects. 

% \textbf{4) Data engine}: The enhanced classification capabilities of Alpha-CLIP when combined with masks/regions fortify its potency as a superior dataset annotator, especially when integrated with SAM~\cite{SAM}.  The data generation pipeline exhibits a conspicuous advantage for \textbf{open vocabulary detection} task compared to the use of the original CLIP.

In summary, we propose Alpha-CLIP, which equips the original CLIP model with the capability of region awareness. Through fine-tuning on millions of RGBA region-text pairs, Alpha-CLIP demonstrates significant advantages over the original CLIP across various tasks, including but not limited to image recognition~\cite{CLIP,ReCLIP,Detic}, multimodal large language models~\cite{BLIP-2,LLaVA}, 2D generation~\cite{dalle2,BLIP-Diffusion} and 3D generation~\cite{Point-E,PureCLIPNeRF}. \vspace{-1mm}

\begin{figure*}[t]
    \centering
    \includegraphics[width=\linewidth]{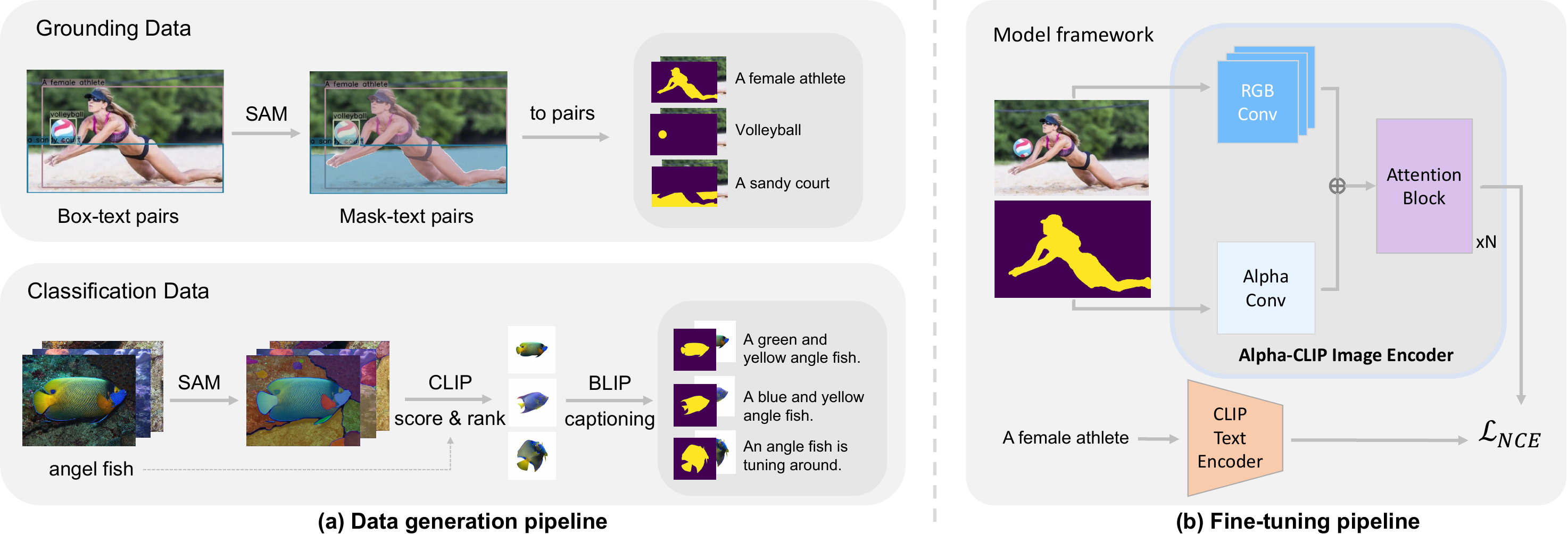}
    \vspace{-7mm}
    \caption{\textbf{The pipeline of our data generation method and model architecture}.
    \textbf{(a)} Our method generates millions of RGBA-region text pairs. \textbf{(b)} \methodname modifies the CLIP image encoder to take an additional alpha channel along with RGB.
    % \shu{You need to unify the terminology, e.g., ``Alpha CLIP'' vs. ``Alpha-CLIP''. Here uses the former whereas the title uses the latter.}
    }
    \vspace{-4.5mm}
    \label{fig:overall}
\end{figure*}

\section{Related Work}

\vspace{-0.5mm}
\noindent \textbf{Empowering CLIP with region awareness.} 
% \noindent \textbf{Works done to make CLIP focus on the specific region.} 
To enable CLIP~\cite{CLIP} to disentangle regions from the whole image for more targeted processing and understanding, various methods have been explored in the field of \textit{segmentation}. Among them, MaskCLIP~\cite{MaskCLIP} uses a 1x1 convolution layer to extract CLIP's final 2D features
% instead of the attention to the original full-image feature
to obtain semantic information for different regions. SAN~\cite{SAN} trains a side network alongside CLIP to assist the model in local semantic perception. MaskCLIP~\cite{attention_mask} and ODISE~\cite{ODISE} use attention masks to make CLIP focus more on local regions. These methods do not alter the weights of the CLIP model itself. RegionCLIP~\cite{zhong2022regionclip} generate region box-text pairs for local region and fine-tune CLIP model for box level recognition. MaskAdaptedCLIP\cite{MaskAdaptedCLIP} generates mask-text pairs for local masks through a pseudo-labeling process and fine-tunes the CLIP model to make it more adaptable to masked images. MaskQCLIP\cite{maskQCLIP} fine-tunes attention layer for new mask \texttt{[CLS]} tokens to make it more fit for mask object classification. These two methods attempt to enhance CLIP's ability to focus on local features and exclusively fine-tune CLIP on specific downstream datasets, resulting in poor generalization ability beyond detection or segmentation tasks.

Another approach is to \textit{change the input image} by simply cropping or masking the image to leave only the foreground object. ReCLIP~\cite{ReCLIP} and OvarNet~\cite{OvarNet} crop the original image using bounding box from object proposal network~\cite{mattnet} and are applied on Referring Expression Comprehension and Open Attribute Recognition tasks. MaskAdaptedCLIP~\cite{MaskAdaptedCLIP} sets the background area to pure color in pixel space and uses the masked image as  input for open-vocabulary segmentation. However, the valuable context information is lost except for using complex post-process proposed in ReCLIP~\cite{ReCLIP}. Some other approaches prompt the CLIP by modifying the input image, guiding CLIP to focus on the area of interest. 
For example, Red-Circle~\cite{circleCLIP}, FGVP~\cite{yang2023fine} use a circle or mask contour to tell CLIP where to focus. Overall, the quality of these approaches that \textit{change the original content of input image} is heavily contingent upon the symbols in CLIP's pre-training dataset. Another limitation is directing modification of images causes a domain gap with CLIP pertaining images.
Unlike previous approaches that rely on \textit{segmentation} or \textit{changing the input image}, our \methodname incorporates an additional alpha channel, which does not change the image content and preserves the generalization performance (cf. \cref{fig:region_image_generation}). \vspace{-0.5mm}

\noindent \textbf{Region-level image  annotation.}
% Existing CLIP models are pretrained on large-scale datasets like LAION-400M~\cite{Laion-400m} or even LAION-5B~\cite{Laion-5b}, while datasets with fine-grained mask-level labels cannot reach such scale due to high manual labor costs. Recently, Kosmos-2~\cite{Kosmos-2} introduced a method that leverages the open detection model GLIP~\cite{GLIP} to transform image-text pairs into more fine-grained pseudo-labels of local image boxes and their corresponding expressions. This dataset, GRIT is used to train multimodal model~\cite{kosmos-1}, equipping it with local perception capabilities and can do referring expression generation. \cite{wang2023allseeing} generate more fine grained text label including attributes and QA-dialog.  Meanwhile, SA~\cite{SAM} has trained a segmentation model SAM on purely vision dataset, enabling it with strong zero-shot abilities for tasks like box-to-mask conversion and automatic mask generation. These developments have made it possible to generate pseudo-masks with region caption at a large scale and have opened up the potential for greater adjustments to CLIP for region level recognition. Therefore, We builds upon GRIT\cite{Kosmos-2} and SAM~\cite{SAM} to propose a method for generating RGBA region-text pairs from grounding data. 
Existing CLIP models are pretrained on large-scale datasets like LAION-400M~\cite{Laion-400m} and  LAION-5B~\cite{Laion-5b}, while fine-grained mask-level labels are not available due to high manual labor costs. Recently, Kosmos-2~\cite{Kosmos-2} introduced a pseudo-labeling pipeline that uses the pre-trained GLIP~\cite{GLIP} model to automatically generate fine-grained pseudo-labels of region boxes and their associate expressions. By using this pseudo-labeling baseline, Kosmos-2 releases the GRIT dataset and equips multimodal model~\cite{kosmos-1} with local perception capabilities. Similarly, the All-Seeing~\cite{wang2023allseeing} project also generates fine-grained text labels via the pseudo-labeling pipeline. Meanwhile, the recent SAM~\cite{SAM} model is trained on massive vision modality data with strong zero-shot abilities for downstream tasks like box-to-mask conversion and automatic mask generation. These developments have made it possible to generate pseudo-masks with region captions at a large scale and have opened up the potential for greater adjustments to CLIP for region-level recognition. Therefore, We build upon GRIT\cite{Kosmos-2} and SAM~\cite{SAM} to propose a method for generating RGBA region-text pairs from grounding data. \vspace{-0.5mm}

\noindent \textbf{CLIP in MLLM.}
% At the age of MLLMs~\cite{flamingo,bai2022ofasys,BLIP-2,palm,gpt4,LLaVA,otter,Kosmos-2,qwen,llava-1.5}. CLIP~\cite{CLIP} has been chosen to serve as vision backbone in the main stream for its semantic representative feature and promising scalability. To make MLLM focus on specific region, Kosmos-2~\cite{Kosmos-2} use millions of region-caption data to train the model to focus on specified region given by box corner points. GPT4ROI~\cite{zhang2023gpt4roi} and GLaMM~\cite{rasheed2023glamm} propose to do ROI pooling on CLIP image feature to refer specific region. These methods only support region focus given at box or patch feature level and training with CLIP vision encoder fixed while our work achieve mask level region prompt and achieve region focusing ability on CLIP model directly.
At the age of Multi-modal Large Language Models~(MLLMs)~\cite{flamingo,bai2022ofasys,BLIP-2,palm,gpt4,LLaVA,otter,kosmos-1,Kosmos-2,qwen,llava-1.5,zhang2023internlm}, CLIP~\cite{CLIP} has been widely used as the vision backbone for its semantic representative feature and promising scalability. To make MLLM focus on the specific region, Kosmos-2~\cite{Kosmos-2} uses millions of region-caption data to train the model with the guidance of box corner points. GPT4ROI~\cite{zhang2023gpt4roi} propose to apply the ROI Align~\cite{he2017mask} operator on the CLIP image feature to refer to the specific region. GLaMM~\cite{rasheed2023glamm} further adds an extra region encoder. Different from previous methods that only support box-level focusing and rely on training additional networks, our work achieves more fine-grained mask-level region focusing and merely uses the CLIP model. \vspace{-0.25mm}

\noindent \textbf{CLIP in 2D image variation.}
% 2D image variation with CLIP image encoder is pioneered by~\cite{dalle2}. Works done in~\cite{diffusers,ip-adapter} also use CLIP to encode image for image variation to achieve better quality or controllability. Subject-driven image variation pioneered by dreambooth~\cite{dreambooth} generate same subject in different background. Following method~\cite{wei2023elite} proposes to use feature level masking to eliminate background information to generate better subject while BLIP-Diffusion~\cite{BLIP-Diffusion} uses text to extract most relevant object feature. All these subject driven image variation methods require the image having a single foreground object in the center of the image and cannot achieve variation focusing on user specified object in more complex image while maintaining original context information.
CLIP image encoder is widely used in 2D image variation (\eg, DALLE-2~\cite{dalle2}, Diffusers~\cite{diffusers} and IP-Adapter~\cite{ip-adapter}) to achieve better quality or controllability. As for subject-driven image variation pioneered by DreamBooth~\cite{dreambooth}, extraction of pure single object feature from the whole image is more important as the following method ELITE~\cite{wei2023elite} proposes to use feature-level masking to eliminate background information to generate better subjects. Similarly, BLIP-Diffusion~\cite{BLIP-Diffusion} uses text to extract the most relevant object features. All these subject-driven image variation methods require the image to have a single foreground object in the center of the image and cannot achieve variation by focusing on user-specified objects in more complex images while maintaining original context information.
Such limitations highlight the importance of our \methodname that enables subject-driven generation in complex scenes and achieves user-defined region focusing in image variation tasks. \vspace{-0.25mm}

\noindent \textbf{CLIP in 3D generation.}
% There are different 3D object generation methods that involve CLIP~\cite{CLIP} model. Two main approaches are based on diffusion model and Neural field optimization respectively. The first approach, pioneered by Point-E~\cite{Point-E}, uses point cloud diffusion model to generate point cloud directly conditioned by CLIP feature from single view image or text. The second approach for 3D object generation involves rendering 2D images from different viewpoints of the neural field and using a 2D vision backbone to guide the optimization direction. One common choice for the vision backbone is CLIP. CLIP loss is first introduced in Dream Fields~\cite{dream_fields} and is used in many following works in text-to-3D including PureCLIPNeRF~\cite{PureCLIPNeRF}, CLIP-Mesh~\cite{CLIP-Mesh}, CLIP-Forge~\cite{CLIP-Forge} and Dream3D~\cite{Dream3D}.
Some existing 3D object generation methods involve CLIP~\cite{CLIP} model. 
In diffusion based 3D generation, Point-E~\cite{Point-E} uses the point cloud diffusion model to generate the point cloud directly conditioned by the CLIP feature from a single view image or text. Another approach in the field of text-to-3D is pioneered by Dream Fields~\cite{dream_fields}, which uses the CLIP model to provide supervision loss. Following works include PureCLIPNeRF~\cite{PureCLIPNeRF}, CLIP-Mesh~\cite{CLIP-Mesh}, CLIP-Forge~\cite{CLIP-Forge} and Dream3D~\cite{Dream3D} also use CLIP image encoder to extract rendered image features.
Our \methodname can enhance CLIP in 3D object generation, enable Point-E with user-defined region focus ability and help optimization based text-to-3D models to yield high-quality generation results.

\section{Method}
This section describes the data pipeline and framework of \methodname. As illustrated in Fig.~\ref{fig:overall}, we first design a data pipeline to generate RGBA-region text pairs data (Sec.~\ref{sec:method_data_pipeline}). Using our generated data, we then train our \methodname with additional Alpha-channel inputs (Sec.~\ref{sec:method_framework}).

\subsection{RGBA Region-Text Pair Generation}
% \noindent In order to finetune the CLIP model, enabling it to accommodate the additional input of an alpha channel, we have designed a data generation pipeline shown in the left part of \cref{fig:overall} for creating RGBA-region text pairs. This pipeline is used to generate millions of data pairs for training CLIP and consists of the following two main components:
To fine-tune the CLIP model with an additional alpha channel input, we first design a data generation pipeline~(cf. \cref{fig:overall}a) to create millions of RGBA-region text pairs. Our d pipeline consists of the following two components.
~\label{sec:method_data_pipeline}
% \noindent \textbf{Grounding Data Pipeline} As depicted in the upper part of \cref{fig:overall}, this branch is dedicated to generating region-text pairs, which include natural images with foreground alpha channels and corresponding referring expressions for specific regions. To be specific, we make use of the GRIT dataset proposed by Kosmos-2~\cite{Kosmos-2}. This dataset employs GLIP to propose matches between proposals and noun referring expressions in image-text pairs, thus automatically labeling them. Furthermore, CLIP is used to score cropped images based on their semantic similarity, allowing high-quality results to be retained. Building upon the GRIT dataset, we use SAM~\cite{SAM} to automatically generate pseudo-masks of high equality.

\noindent \textbf{Grounding data pipeline.} As depicted in the upper part of \cref{fig:overall}a, this branch is dedicated to generating region-text pairs, which include natural images with foreground alpha channels and corresponding referring expressions for specific regions. The natural images are from the GRIT dataset~\cite{Kosmos-2}, which employs GLIP and CLIP to automatically extract labels of \textit{box} region-text pairs. Building upon GRIT, we take a further step of generating \textit{mask} region-text pairs. Specifically, we use SAM~\cite{SAM} to automatically generate high-equality pseudo-masks for each box region.

\noindent \textbf{Classification data pipeline.}
\label{para:data_pipline_object_centric}
As illustrated in the lower part of \cref{fig:overall}a, this branch is utilized for generating region-text pairs where the foreground objects are highlighted while the original background is removed. We employ the ImageNet~\cite{Imagenet21k} dataset for this purpose. Firstly, we use SAM to automatically generate several masks for each image in ImageNet. Subsequently, we crop the foreground object of each mask, center it, and enlarge it. CLIP is then used to calculate scores with the corresponding class label of the image to which each mask belongs. Following this, we sort the masks by class based on their scores and select the top-ranked masks with the highest scores. Regarding the text component, to ensure that the caption for each mask is not merely the ImageNet~\cite{Imagenet21k} class label, we place the foreground object on a pure white background. Then we use BLIP-2~\cite{BLIP-2} to annotate these masks with captions. Finally, we merge the fine-grained ImageNet class label with the image-specific captions generated by BLIP-2~\cite{BLIP-2}, resulting in millions of RGBA region-text pairs.

\subsection{Alpha-CLIP}
% \noindent\textbf{Model structure.} To maximize the utilization of CLIP's prior knowledge, we made subtle structural adjustments to the CLIP-Vision Encoder. Specifically, in the CLIP-Vision's ViT~\cite{ViT} structure, a large kernel size convolution is applied to the image in the first layer. As shown in the right part of \cref{fig:overall}. To allow the CLIP-Vision Encoder to accept an additional alpha channel as input, we add an additional Alpha Conv layer parallel to the RGB Conv layer and take input defined to range from $[0,1]$ (1 represents the foreground and 0 represents the background). Before training, we initialize Alpha Conv kernel weights to all zeros, ensuring that the initial Alpha-CLIP ignores the alpha channel as input.

\noindent\textbf{Model structure.} Our \methodname implements subtle structural modifications to the CLIP image encoder to preserve CLIP's prior knowledge. In the CLIP image encoder's ViT~\cite{ViT} structure, an RGB convolution is applied to the image in the first layer. As shown in \cref{fig:overall}b, we introduce an additional Alpha Conv layer parallel to the RGB Conv layer, which enables the CLIP image encoder to accept an extra alpha channel as input. The alpha channel input is set to range from $[0,1]$, where 1 represents the foreground and 0 indicates the background. We initialize the Alpha Conv kernel weights to zero, ensuring that the initial \methodname ignores the alpha channel as input. \vspace{1mm}

\noindent \textbf{Training method.} During training, we keep the CLIP text encoder fixed and entirely train the \methodname image encoder. Compared to the first convolution layer that processes the alpha channel input, we apply a lower learning rate to the subsequent transformer blocks. To preserve CLIP's global recognition capability for full images, we adopt a specific data sampling strategy during training. We set the sample ratio, denoted as $r_s=0.1$ to occasionally replace our generated RGBA-text pairs with the original image-text pairs and set the alpha channel to full 1. Please refer to~\cref{app_hyperparameters} for ablation studies such as the number of unfreeze Transformer blocks and value of $r_s$. \vspace{1mm}

\noindent \textbf{Alpha-CLIP for downstream tasks.}
% \noindent After the training as described above, Alpha-CLIP possesses the capability to focus on a specified region and generate corresponding features. Given the widespread use of CLIP in various tasks, Alpha-CLIP has a wide range of application scenarios. We illustrate the advantages of Alpha-CLIP over the original CLIP~\cite{CLIP} by showcasing its use in various tasks as listed in \cref{tab:tasks} in Sec.~\ref{sec:introduction}.
After the training, Alpha-CLIP possesses the capability to focus on a specified region and controlled editing. Alpha-CLIP can enhance CLIP's performance on various baselines in a plug-and-play fashion, across various downstream tasks like recognition, MLLM, and 2D/3D generation (see \cref{tab:tasks} in \cref{sec:introduction}).
~\label{sec:method_framework}

\section{Experiments}
% \noindent We train Alpha-CLIP on RGBA region-text pairs using grounding data pipeline from GRIT~\cite{Kosmos-2}-20m for zero-shot ImageNet classification. And combined it with 450k RGBA region-text pair from ImageNet~\cite{Imagenet21k} using classification data pipeline to train Alpha-CLIP for other tasks including REC, data engine for OVD, region level captioning, 2D image variation and 3D generation.
\noindent \textbf{Data.} We train \methodname on RGBA region-text pairs using grounding data pipeline from GRIT-20m~\cite{Kosmos-2} for zero-shot ImageNet classification. We combine it with 460k RGBA region-text pair from ImageNet~\cite{Imagenet21k} using classification data pipeline to train Alpha-CLIP for other tasks including REC, OVD, region-level captioning, 2D image variation, and 3D generation. Ablation on data volume and mixture of data are in \cref{app_ablation_data_volume,app_ablation_mixture_of_data}
\subsection{\methodname in Image Recognition}
 % \vspace{-2mm}
{
\setlength{\tabcolsep}{3.4mm}
\begin{table}[t]
  \centering
  \vspace{-2mm}
  \scalebox{0.80}{
%     \begin{tabular}{cc|ccc}
%     \toprule
%     \multicolumn{2}{c}{Model} & ViT-B/16 & ViT-L/14 \\
%     \midrule
%     \multicolumn{2}{c|}{Original CLIP} & 66.48 / 88.90  & 73.48 / 91.60 \\
%     \multicolumn{2}{c|}{MaskAdaptedCLIP~\cite{MaskAdaptedCLIP}} & 57.86 / 79.12  & 63.50 / 86.34  \\
% \multicolumn{2}{c|}{CircleCLIP~\cite{circleCLIP}} & 65.37 / 88.68  & 73.37 / 92.09  \\
% \multicolumn{2}{c|}{MaskCLIP*~\cite{MaskCLIP}} & 67.86 / 89.40  & 77.04 / 93.39  \\

%     \rowcolor{violet!10} \multirow{2}[0]{*}{ Ours} & GRIT-1m & \bfseries{68.30} / \bfseries{90.31} & \bfseries{77.22} / \bfseries{94.38} \\
%     & GRIT-20m & \bfseries{68.89} / \bfseries{90.51} & \bfseries{77.58} /  \bfseries{94.47}\\
%     \bottomrule
%     \end{tabular}%
    \begin{tabular}{c|cc|cc}
    \toprule
    \multirow{2}{*}{\textbf{Methods}} & \multicolumn{2}{c|}{ViT-B/16} & \multicolumn{2}{c}{ViT-L/14} \\
    & Top-1 & Top-5 & Top-1 & Top-5 \\
    \midrule
    Original CLIP~\cite{CLIP} & 66.48 & 88.90  & 73.48 & 91.60 \\
    MaskAdaptedCLIP~\cite{MaskAdaptedCLIP} & 57.86 & 79.12  & 63.50 & 86.34  \\
    Red Circle~\cite{circleCLIP} & 65.37 & 88.68  & 73.37 & 92.09  \\
    MaskCLIP*~\cite{MaskCLIP} & 67.86 & 89.40  & 77.04 & 93.39  \\
    \rowcolor{violet!10} Alpha-CLIP(ours) & \bfseries{68.89} & \bfseries{90.51} & \bfseries{77.41} &  \bfseries{94.45}\\
    \bottomrule
    \end{tabular}%
    }
    \vspace{-3mm}
    \caption{\textbf{Zero-shot classification on ImageNet-S~\cite{gao2022large}.}
  % After training Alpha-CLIP with millions of RGBA-region text pair. 
  % When given foreground object on alpha channel, our Alpha-CLIP significantly improves zero-shot classification and surpasses MaskCLIP~\cite{MaskCLIP}. Note MaskCLIP is not designed for recognition but mask generation; we make necessary modifications on MaskCLIP to improve recognition.
  When given the foreground object on the alpha channel, our \methodname significantly improves zero-shot classification and surpasses previous baselines such as MaskCLIP~\cite{MaskCLIP}.
  % \shu{Are there other baselines? For example, there should be a baseline method that directly masks out RGB regions as input to CLIP. How about MaskCLIP?}
  % \shu{Are there more SOTA methods other than MaskCLIP?}
  % \shu{can reformat this table to list top-1 and top-5 nubmers like this: xx.xx / yy.yy. This makes comparison easier and saves space.}
  \vspace{-1mm}
  }
  \label{tab:imagenet_table}%
\end{table}%
}

{
\setlength{\tabcolsep}{3.4mm}
\begin{table}[t]
  \centering
  % \vspace{-2mm}
  \scalebox{0.89}{
    \begin{tabular}{c|c|cc}
    \toprule
    \textbf{Model} & \textbf{Alpha Map} & \textbf{Top-1} & \textbf{Top-5} \\
    \midrule
    CLIP~\cite{CLIP} & - & 73.48 & 91.60 \\
    \multirow{3}[0]{*}{Alpha-CLIP} & \cellcolor{violet!5} whole image & \cellcolor{violet!5} 73.37  & \cellcolor{violet!5} 91.75  \\
    & \cellcolor{violet!10} rectangular box & \cellcolor{violet!10} 75.62  & \cellcolor{violet!10} 93.34  \\
    & \cellcolor{violet!15} mask & \cellcolor{violet!15} 77.41  & \cellcolor{violet!15} 94.45  \\
    \bottomrule
    \end{tabular}%
    }
    \vspace{-2mm}
    \caption{
    % \textbf{Zero-shot classification on ImageNet-S\cite{gao2022large} when different alpha map is available} Alpha-CLIP classification accuracy when given different level alpha map. When there is no focused region available, Alpha-CLIP can still maintain original CLIP recognition ability.
    \textbf{Zero-shot classification on ImageNet-S~\cite{gao2022large} with different alpha map levels.} \methodname is comparable to the original CLIP when the foreground mask is not available, and further boosts the performance with rectangular box or mask alpha maps.
  }
  \vspace{-4mm}
  \label{tab:imagenet_diff_prompt}%
\end{table}%
}

\noindent \textbf{Zero-shot image classification.}
% To test the effectiveness of Alpha-CLIP, we use ImageNet-S~\cite{gao2022large} dataset. This dataset comprises 919 classes selected from ImageNet-1k and includes semantic segmentation annotations for the images belonging to these 919 classes. We prepare the semantic segmentation masks of the image level label as the alpha channel input. 
We select the ImageNet-S~\cite{gao2022large} dataset for zero-shot classification analysis, which comprises 919 classes with semantic segmentation annotations selected from ImageNet-1k. We prepare the image-level semantic segmentation masks as the alpha channel input. We select representative baseline methods designed for making CLIP focus on the specific region: MaskCLIP~\cite{MaskCLIP}, MaskAdaptedCLIP~\cite{MaskAdaptedCLIP}, and Red Circle~\cite{circleCLIP}. Note that MaskCLIP is designed for mask generation rather than recognition. We make necessary modifications to MaskCLIP to adapt it for the recognition task (please refer to \cref{app_MaskCLIP} for our implementation details).
% To ensure a fair comparison, we also fine-tune the original CLIP directly with the original image-text pairs as one baseline method.
We use the mean of per-class accuracy as the evaluation metric. 

% We evaluate Alpha-CLIP's zero-shot classification performance on the validation set of ImageNet-S, using mean of per-class accuracy as the evaluation metric. We also compare the result with MaskCLIP~\cite{MaskCLIP}(implementation details in \cref{app_MaskCLIP}), MaskAdaptedCLIP~\cite{MaskAdaptedCLIP} and Red Circle~\cite{circleCLIP}. Results are shown in \cref{tab:imagenet_table}. To ensure a fair comparison, we also fine-tune the original CLIP directly with original image-text pair, but only observe negligable improvement. This experiment effectively demonstrates that after training Alpha-CLIP with millions of RGBA-region text pairs, when provided with a foreground object mask through alpha channel, our Alpha-CLIP can generate visual feature that is more focused on the foreground object, leading to better image-level classification compared to the original CLIP. It is worth noticing that although MaskCLIP~\cite{MaskCLIP} can achieve pretty good result without fine-tuning CLIP model. it can not be directly used to methods where the whole feature map instead of \texttt{[CLS]} token only is required, like BLIP-2\cite{BLIP-2}, BLIP-Diffusion~\cite{BLIP-Diffusion}, LLaVA~\cite{LLaVA} and Point-E~\cite{Point-E}.
\cref{tab:imagenet_table} presents the zero-shot classification comparison on ImageNet-S \textit{validation} set. This experiment effectively demonstrates that when provided with a foreground object mask through the alpha channel, our \methodname generates visual features that are more focused on the foreground object, leading to better image-level classification compared to the original CLIP and other baseline approaches.
It is worth noticing that Although MaskCLIP~\cite{MaskCLIP} achieves good results without needing to fine-tune the CLIP model, it is not directly compatible with methods that require the whole feature map instead of just the \texttt{[CLS]} token. This limitation is particularly relevant when considering methods like BLIP-2\cite{BLIP-2}, BLIP-Diffusion~\cite{BLIP-Diffusion}, LLaVA~\cite{LLaVA} and Point-E~\cite{Point-E}. In contrast, our \methodname is more general and can be applied to these approaches effectively.

% We also test Alpha-CLIP when foreground mask is not available. As shown in \cref{tab:imagenet_diff_prompt}, when foreground prior is not available, we set alpha channel input to all one, in which Alpha-CLIP can still maintain the original CLIP recognition ability. When provided foreground box or foreground mask, Alpha-CLIP can significantly improve classification accuracy.
We also evaluate Alpha-CLIP in scenarios where the foreground mask is unavailable. As shown in \cref{tab:imagenet_diff_prompt}, when foreground prior is not available, we set alpha channel input to all one. We observe that the recognition ability of Alpha-CLIP (second row) remains on par with the original CLIP (top row). When provided foreground box (third row) or foreground mask (bottom row), Alpha-CLIP can significantly improve classification accuracy. \vspace{1mm}

\noindent \textbf{Zero-shot referring expression comprehension.}
% In addition to zero-shot image classification testing, we conducted experiments on zero-shot Referring Expression Comprehension (REC) to test the model's inherent image-text understanding and reasoning capabilities. This involved the task of localizing objects in an image given a textual reference expression in a zero-shot manner. REC is commonly evaluated on the RefCOCO~\cite{refcoco}, RefCOCO+~\cite{refcoco}, and RefCOCOg~\cite{refcocog} datasets.
In addition to the zero-shot image classification task, we also conducted experiments on zero-shot Referring Expression Comprehension (REC). zero-shot REC is the task of localizing objects in an image given a textual reference expression in a zero-shot manner. We follow previous works to select the RefCOCO~\cite{refcoco}, RefCOCO+~\cite{refcoco}, and RefCOCOg~\cite{refcocog} datasets for evaluation.
% RefCOCO+ only contains appearance-based expressions, whereas RefCOCO and RefCOCOg contain relation-based expressions (e.g., containing the words right/closer/smaller). 
% Following prior work~\cite{ReCLIP, circleCLIP, cpt}, we utilize object proposals predicted by a pretrained detector, following a similar paradigm as employed in MAttNet~\cite{mattnet}. We employed SAM to obtain masks for each proposal, replacing the CLIP model in ~\cite{ReCLIP} with Alpha-CLIP. Additionally, we remove the image cropping preprocessing method, and instead input both the original image and mask into Alpha-CLIP, which helps to retain valuable global contextual information when directing the model's attention to the region.
We select three representative approaches CPT~\cite{cpt}, ReCLIP~\cite{ReCLIP}, and Red-Circle~\cite{circleCLIP} as our baselines. We replace the CLIP model in this task with our \methodname. Specifically, we use object proposals predicted by a pretrained detector~\cite{mattnet} and employ SAM to obtain masks for each proposal. Instead of cropping the object by bounding box, we input the original image with an alpha map into our Alpha-CLIP. This modification has proven beneficial in preserving global contextual information as we find cropping only lead to worse result. Please refer to \cref{sec:app_zero_shot_REC_imp} for more implementation details.

% As shown in \cref{tab:ref-exp}, Alpha-CLIP achieves competitive zero-shot results on the REC task, surpassing ReCLIP and RedCircle by an average of 6.8\% and 3.0\% accuracy across RefCOCO, RefCOCO+, and RefCOCOg benchmarks.
As shown in \cref{tab:ref-exp}, Alpha-CLIP achieves competitive zero-shot results on the REC task, surpassing ReCLIP and RedCircle by an average of 6.8\% and 3.0\% accuracy across RefCOCO, RefCOCO+ and RefCOCOg benchmarks. The experimental results demonstrate that \methodname enhances CLIP's ability to focus on the relevant region and such enhancement is also beneficial for the REC task that requires image-text understanding and reasoning capabilities. \vspace{1mm}

\begin{table}
% \vspace{-4mm}
\scalebox{0.70}{
\begin{tabular}{lccc|ccc|cc}
    \toprule
\multirow{2}{*}{\textbf{Method}} &\multicolumn{3}{c|}{\textbf{RefCOCO}} & \multicolumn{3}{c|}{\textbf{RefCOCO+}} & \multicolumn{2}{c}{\textbf{RefCOCOg}} \\
 & Val & TestA & TestB & Val & TestA & TestB & Val & Test \\
\midrule
CPT~\cite{cpt} & 32.2 & 36.1 & 30.3 & 31.9 & 35.2 & 28.8 &  36.7 & 36.5 \\
ReCLIP~\cite{ReCLIP} & 45.8 & 46.1 & 47.1 & 47.9 & 50.1 & 45.1 & 59.3 & 59.0 \\
Red Circle~\cite{circleCLIP} & 49.8 & 58.6 & 39.9 & 55.3 & \textbf{63.9} & 45.4 & 59.4 & 58.9 \\
\rowcolor{violet!10} Alpha-CLIP & \textbf{55.7} & \textbf{61.1} & \textbf{50.3} & \textbf{55.6} & 62.7 & \textbf{46.4} & \textbf{61.2} & \textbf{62.0} \\
    \bottomrule \\
\end{tabular}
}
\vspace{-6mm}
\caption{\textbf{Comparison with state-of-the-art on zero-shot REC.} We report top-1 accuracy (\%). Replacing CLIP in ReCLILP~\cite{ReCLIP} with Alpha-CLIP outperforms other zero-shot approaches on most datasets, including Red Circle\cite{circleCLIP}, ReCLIP\cite{ReCLIP} and CPT\cite{cpt}.}
\vspace{-6mm}
% \vspace{-2em}
%
\label{tab:ref-exp}
\end{table}

\noindent \textbf{Open vocabulary detection.}
% Detic~\cite{Detic} proposes the use of the ImageNet dataset to assist in OVD tasks. Specifically, in the first training round, only the LVIS~\cite{LVIS} dataset is utilized. In the second round, leveraging the detector's existing detection capabilities, bounding boxes are employed to pseudo-annotate the ImageNet dataset, which only contains image-level labels. This enhancement aims to boost the classifier's performance in OVD tasks. By following the methodology outlined in \cref{para:data_pipline_object_centric}, the ImageNet dataset was transformed into a pseudo-labeled detection dataset with foreground masks. The ImageNet data used in the second round of Detic's training was replaced with this MaskImageNet dataset. Background category loss is removed, and the blending ratios of LVIS and MaskImageNet is adjusted. Experimental results, as shown in the table, demonstrate that utilizing the raw CLIP method for labeling already enhances OVD capabilities. Furthermore, employing the Alpha-CLIP labeling method further improves OVD performance. Remarkably, while Detic utilized 1200k images from ImageNet in the OV-LVIS~\cite{LVIS} benchmark, we achieved further improvements in OVD capabilities using only top ranked 450k images.
The Open-Vocabulary Detection (OVD) task aims to detect novel classes that are not available during training.
Detic~\cite{Detic} is a pseudo-labeling baseline that proposes to use the ImageNet dataset for OVD. Specifically, Detic first trains the detector on the base classes of LVIS~\cite{LVIS}, then uses the detector to generate pseudo bounding boxes on ImageNet. These pseudo boxes may cover the novel objects and help improve the detector's performance in novel classes. Such a semi-supervised pipeline is not data-efficient and Detic uses 1.2M images from ImageNet in the OV-LVIS~\cite{LVIS} benchmark.

To demonstrate the effectiveness of \methodname on OVD, we transfer the top-ranked ImageNet (460K) into a collection, dubbed as MaskImageNet. Specifically, we apply our data generation pipeline, as detailed in \cref{para:data_pipline_object_centric} to generate pseudo-labeled bounding boxes and foreground masks for each image. We replace the ImageNet used in Detic's pseudo-labeling steps with our MaskImageNet. We also remove the background category loss and adjust the blending ratios of LVIS and MaskImageNet. Experimental results are presented in ~\cref{tab:addlabel}. Compared to the Detic baseline using ImageNet (top row), The second row demonstrates that using our MaskImageNet already enhances OVD capabilities. Furthermore, our Alpha-CLIP (bottom row) further improves OVD performance. Remarkably, our method (460K in MaskImageNet) is more data efficient than Detic (1.2M in ImageNet).

{

\begin{table}[t]
  \centering
  \vspace{-2mm}
  \scalebox{0.9}{
    \begin{tabular}{ccc}
    \toprule
    Dataset  & mAP\scriptsize{novel}   & mAP \\
    \midrule
    Detic-ImageNet & 24.6  & 32.4 \\
    MaskImageNet (ori CLIP) & \bfseries{27.9}  & \bfseries{32.5} \\
    \rowcolor{violet!10} MaskImageNet (Alpha-CLIP) & \bfseries{28.6}  & \bfseries{32.9} \\
    \bottomrule
    \end{tabular}%
    }
    \vspace{-3mm}
    \caption{
    % \textbf{OV-LVIS~\cite{LVIS} test result}. Using MaskImageNet can significantly improve mAP{\scriptsize novel}
    \textbf{Open-vocabulary detection on OV-LVIS~\cite{LVIS}}. Using MaskImageNet and our \methodname can significantly improve mAP{\scriptsize novel} on novel classes.
  }
  \vspace{-7mm}
  \label{tab:addlabel}%
\end{table}%
}

\subsection{Alpha-CLIP in MLLM}
\label{sec:alpha_clip_MLLM}
% \noindent We replace CLIP image encoder used in BLIP2~\cite{BLIP-2} and LLaVA-1.5~\cite{llava-1.5} with Alpha-CLIP to make MLLM directly focus on user defined region in captioning and VQA.
We replace CLIP used in BLIP-2~\cite{BLIP-2} and LLaVA-1.5~\cite{llava-1.5} with our Alpha-CLIP to make MLLM directly focus on user-defined region in vision-language tasks such as region level captioning and VQA.

\begin{figure}[]
    \centering
    \includegraphics[width=\linewidth]{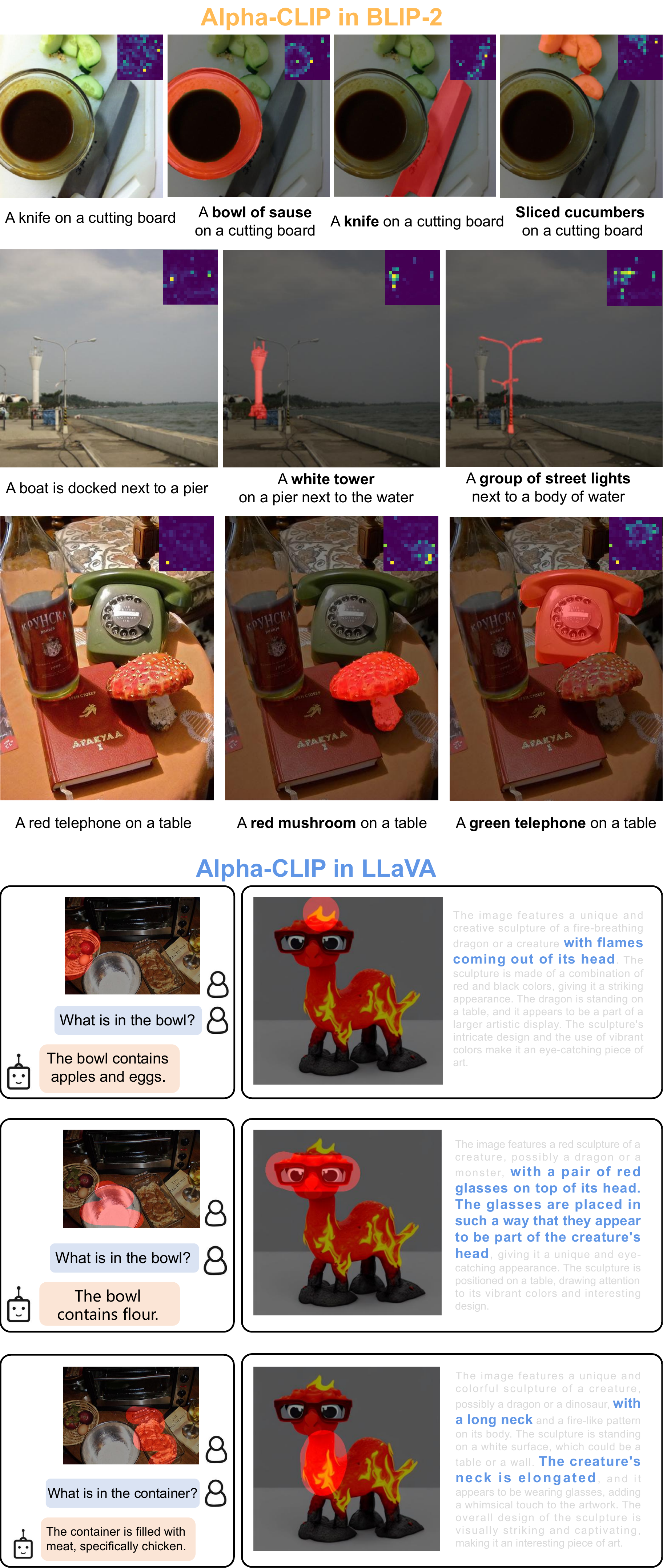}
    \caption{\textbf{Some results of Alpha-CLIP used in MLLMs} The upper half is image captioning result with BLIP-2~\cite{BLIP-2}. The first column is the original CLIP generated captions. Other columns represent the outcomes of Alpha-CLIP with highlighted region marked in red. The lower half is region focused VQA and image captioning result with LLaVA-1.5~\cite{llava-1.5}.}
    \label{fig:figure_w_llm}
\end{figure}

\noindent \textbf{Region level captioning.}
% We use BLIP-2~\cite{BLIP-2} model using CLIP ViT-L/14 with Flan-t5-xl~\cite{Flant} and LLaVA-1.5~\cite{llava-1.5} using CLIP ViT-L/14@336px with Vicuna~\cite{vicuna}. By simply replace CLIP with Alpha-CLIP, we can guide MLLM generate caption more focused on user defined areas. As shown in \cref{fig:figure_w_llm}, the attention map marked on top right corner of the image also verifies the result(more visualizations and implementation detail of attention map are in \cref{sec:more_visualization_attn}) It is worth noticing that in the third demonstrated case about telephone and mushroom, original CLIP with the whole image as input will cause the LLM to generate wrong caption, We suspect that this is because CLIP vision feature mix different objects and their properties when there are too many foreground objects exist in a single image. Prompting with area to focus on, Alpha-CLIP can generate purer feature to guide LLM to generate correct caption. More visualization are in \cref{sec:more_visualization_blip2,sec:more_visualization_llava}.
As shown in \cref{fig:figure_w_llm}, simply replacing CLIP with Alpha-CLIP enables MLLM to generate captions more focused on user-defined areas. In the third row cases about the telephone and mushroom, the original CLIP generates the wrong caption. This error may arise due to the CLIP vision feature mixing different objects and their properties in images with too many foreground objects. \methodname guides MLLM to generate the correct caption by providing the area to focus on. More visualization are in \cref{sec:more_visualization_blip2,sec:more_visualization_llava}. We also visualize the CLIP attention map marked in the upper right to confirm our findings. More visualizations and implementation details of the attention map are in \cref{sec:more_visualization_attn}.

% Besides these qualitative visualization, we also test region level captioning ability of Alpha-CLIP with LLaVA-1.5~\cite{llava-1.5} by quantitative experiment on Visual Genome~\cite{VG} and refCOCOg~\cite{refcocog} following previous work setting. As shown in \cref{tab:region_level_captioning}. After fine-tune on these specific datasets, Alpha-CLIP+LLaVA-1.5 can achieve competitive result on region level captioning task. Even surpass previous expert model like GPT4ROI~\cite{zhang2023gpt4roi} and GLaMM~\cite{rasheed2023glamm} with ROI Align structure and large volume of region-text pairs for pertaining.
Besides qualitative results, we also provide the quantitative region level captioning results of Alpha-CLIP with LLaVA-1.5~\cite{llava-1.5} on Visual Genome~\cite{VG} and RefCOCOg~\cite{refcocog}. We fine-tune Alpha-CLIP+LLaVA-1.5~\cite{llava-1.5} with vicuna-7b~\cite{vicuna} on these datasets with the same setting in \cite{zhang2023gpt4roi,rasheed2023glamm} and task prompts in \cite{zhang2023gpt4roi} is adopted. Alpha-CLIP image encoder is kept frozen with LLM fine-tuned to adapt region caption format.  Results are shown in \cref{tab:region_level_captioning}. Alpha-CLIP+LLaVA-1.5 achieves competitive results over the baseline methods, even surpassing previous expert models like GPT4ROI~\cite{zhang2023gpt4roi} and GLaMM~\cite{rasheed2023glamm} with ROI Align~\cite{he2017mask} or additional region encoder structure pretrained on a large volume of region-text pairs.

\begin{table}[h]
\centering
\scalebox{0.78}{
\begin{tabular}{lcccc}
\toprule
\multirow{2}{*}{\textbf{Model}} & \multicolumn{2}{c}{\textbf{RefCOCOg}} & \multicolumn{2}{c}{\textbf{Visual Genome}} \\ 
\cmidrule(lr){2-3} \cmidrule(lr){4-5} & METEOR & CIDEr & METEOR & CIDEr \\ 
\midrule
GRIT \cite{wu2022grit}& 15.2 & 71.6 & 17.1 & 142.0 \\
Kosmos-2 \cite{Kosmos-2}& 14.1 & 62.3 & - & - \\
GPT4RoI \cite{zhang2023gpt4roi}& - & - & 17.4 & 145.2 \\
% ASM-ZS \cite{wang2023all}& \checkmark & 13.6 & 41.9 & 12.6 & 44.2 \\
% ASM-FT \cite{wang2023all}& \times & 20.8 & 103 & 18.0 & 145.1 \\
GLaMM~\cite{rasheed2023glamm} & 16.2 & 105.0 & 18.6 & 157.8 \\
% \rowcolor{violet!10} 
\rowcolor{violet!10} Alpha-CLIP+LLaVA~\cite{llava-1.5} &\textbf{16.7} & \textbf{109.2} & \textbf{18.9} & \textbf{160.3} \\
\bottomrule
\end{tabular}
}
\vspace{-3mm}
\caption{
% \textbf{Performance of Alpha-CLIP in region level captioning.}: Metrics include METEOR and CIDEr scores, assessed on Visual Genome and refCOCOg Datasets, exhibiting competitive results.
\textbf{Performance of Alpha-CLIP in region level captioning.} We report METEOR and CIDEr metrics on Visual Genome and refCOCOg Datasets.
}
\vspace{-4mm}
\label{tab:region_level_captioning}
\end{table}

\noindent \textbf{Region based VQA.}
% MLLM also have the ability to chat with user and do simple reasoning. In this scenario, alpha channel input can be seen as visual prompt defined by user to refer specify region of interest to talk about. As shown in ~\cref{fig:figure_w_llm} and ~\cref{fig:teaser}, user can simply use stroke to tell MLLM the referring object or regions to focus on. We believe this will have more application when combining with mask proposal network like SAM~\cite{SAM} or OVD model like~\cite{GLIP} More visualization of VQA with Alpha-CLIP are in \cref{sec:more_visualization_llava}
MLLM can chat with users with simple reasoning. In this scenario, alpha channel input can act as the visual prompt defined by the user to highlight specific regions of interest. As shown in \cref{fig:figure_w_llm} and~\cref{fig:teaser}, the user can simply use stroke to tell MLLM the referring object or regions to focus on.
% We believe our approach has more application when combined with the mask proposal network like SAM~\cite{SAM} or OVD model like GLIP~\cite{GLIP}.
More visualization results of VQA with Alpha-CLIP are in \cref{sec:more_visualization_llava}.

\subsection{Alpha-CLIP in image variation.}
\label{sec:alpha_clip_image_var}
% \noindent Alpha-CLIP can be used in most image variation models that use CLIP image encoder~\cite{dalle2, diffusers, wei2023elite, ip-adapter, BLIP-Diffusion}, among which BLIP-Diffusion~\cite{BLIP-Diffusion} is a model that better maintain subject information. So we use this work to demonstrate the effectiveness of Alpha-CLIP. BLIP-Diffusion~\cite{BLIP-Diffusion} model bridges CLIP~\cite{CLIP} and stable-diffusion~\cite{Stable-Diffusion} with Q-former to achieve the ability to generate and edit 2D images controlled by text. By introducing Alpha-CLIP, we can add an additional set of vision prompts to allow the model to focus on specified regions for generation. We replace the ViT-L/14 model in BLIP-Diffusion~\cite{BLIP-Diffusion} with Alpha-CLIP while keeping the other parts unchanged to achieve user-specified region generation.
Alpha-CLIP can be used in most image variation models that use CLIP image encoder~\cite{dalle2, diffusers, wei2023elite, ip-adapter, BLIP-Diffusion}. For example, BLIP-Diffusion bridges CLIP~\cite{CLIP} and stable-diffusion~\cite{Stable-Diffusion} with Q-former to generate and edit 2D images controlled by text. Since BLIP-Diffusion~\cite{BLIP-Diffusion} is a typical method that maintains subject information, we use BLIP-Diffusion to demonstrate the effectiveness of Alpha-CLIP. By introducing Alpha-CLIP, we can add an additional set of vision prompts to allow the model to focus on specified regions for 2D generation. We replace the ViT-L/14 model in BLIP-Diffusion~\cite{BLIP-Diffusion} with Alpha-CLIP while keeping the other parts unchanged.
% Results are shown in \cref{fig:region_image_generation}. 
% We set text prompt to empty to make result irrelevant with semantics. As shown in \cref{fig:teaser}, after replacing CLIP with Alpha-CLIP and prompting it with highlighted areas. it will be more focused on these areas and thus generate purer feature to guild stable-diffusion~\cite{Stable-Diffusion} model to generate new image. We also compare our Alpha-CLIP with other simple modification to make region focused generation by image cropping, pixel level image masking and feature level image masking(implementation details in \cref{sec:masking baseline}). As shown in \cref{fig:region_image_generation}, image cropping can not solve occlusion and Neither feature level nor pixel level masking can convey original background information. While Alpha-CLIP, prompting CLIP with fine grained region can solve above problems and generate cleaner result while maintaining original background information. More visualizations are in \cref{sec:more_visualization_blip-diff}
We set the empty text prompt to make results irrelevant with semantics. As shown in \cref{fig:teaser}, Alpha-CLIP with alpha map on highlighted areas enables BLIP-Diffusion to generate region-focused results. We also compare our Alpha-CLIP with other CLIP region-focused approaches such as image cropping, pixel-level image masking, red circle, and feature-level masking (Please refer to \cref{sec:app_masking_baseline} for implementation details). As shown in \cref{fig:region_image_generation}, image cropping can not solve the occlusion problem. The red-circle solution will change the image content. Neither pixel-level nor feature-level masking can convey original background information. In contrast, our Alpha-CLIP that prompts CLIP with fine-grained region mask solves the above problems and generates cleaner results while maintaining original background information. More visualizations are in \cref{sec:more_visualization_blip-diff}

\subsection{Alpha-CLIP in 3D Object Generation.}
\label{sec:alpha_clip_3d}
% \noindent Alpha-CLIP can also apply to 3D Object Generation. We test it in two different approaches, namely diffusion based Point-E~\cite{Point-E} for image-to-3D and optimization based PureCLIPNeRF~\cite{PureCLIPNeRF} for text-to-3D.
Alpha-CLIP can also apply to 3D Object Generation. We test it in two different approaches: 1) Point-E~\cite{Point-E} that is a diffusion-based method for image-to-3D, and 2) PureCLIPNeRF~\cite{PureCLIPNeRF} that is an optimization-based approach for text-to-3D.

\begin{figure}[t]
    \centering
    \includegraphics[width=\linewidth]{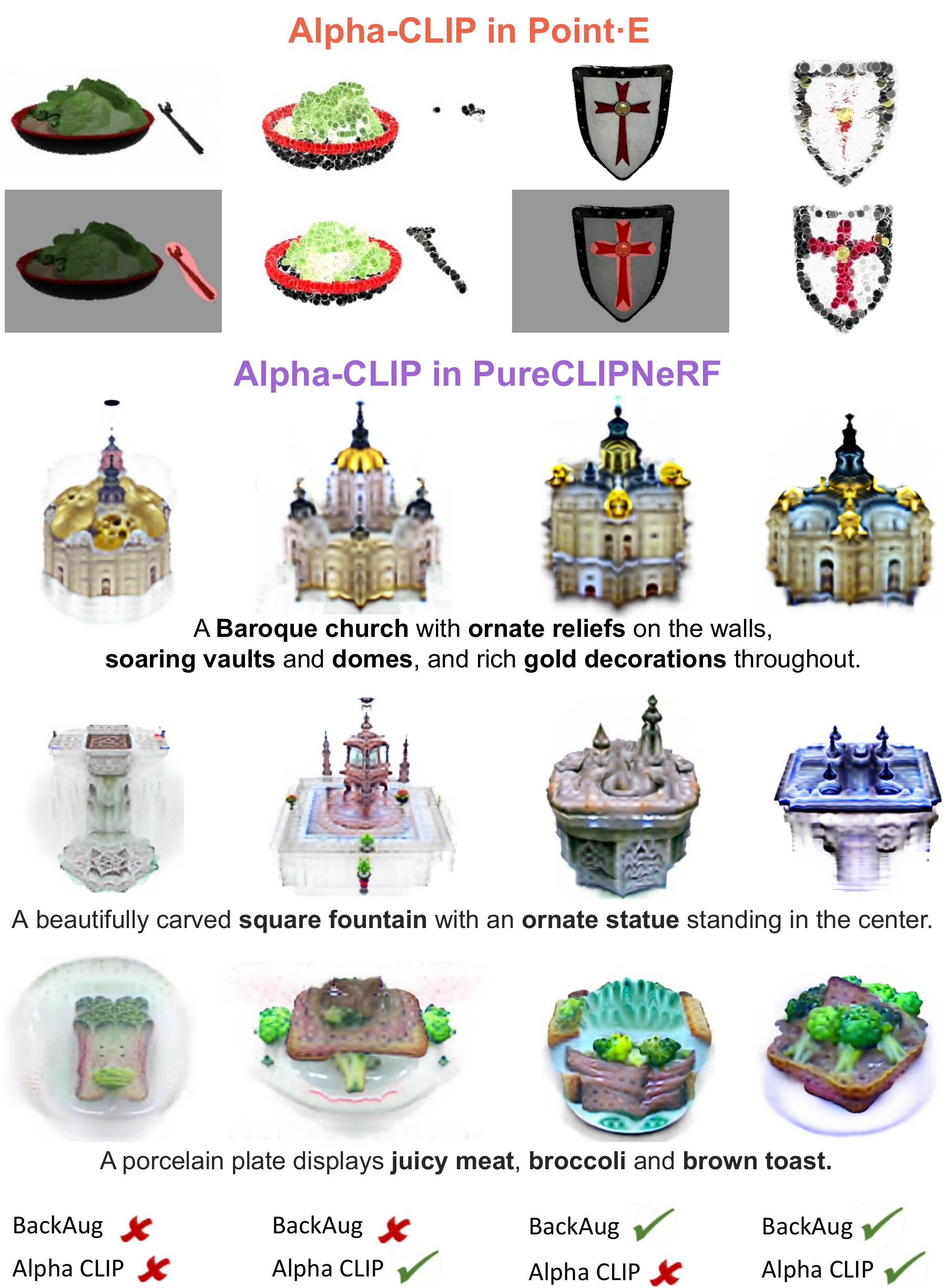}
    \vspace{-5mm}
    \caption{
    % \textbf{Results of Alpha-CLIP in 3D generation.} The top part shows 3D point clouds generation using Point·E~\cite{Point-E}. The first row displays objects generated by the original CLIP. The second row illustrates results of Alpha-CLIP, with highlighted areas in red. The bottom part shows 3D objects generated by PureCLIPNeRF~\cite{PureCLIPNeRF}. The original CLIP model is replaced with Alpha-CLIP, and tests are conducted with and without background augmentation.
    \textbf{Results of Alpha-CLIP in 3D generation.} The top part shows 3D point clouds generation using Point·E~\cite{Point-E}. The first row displays objects generated by the original CLIP. The second row illustrates the results of Alpha-CLIP, with highlighted areas in red. The bottom part shows 3D objects generated by PureCLIPNeRF~\cite{PureCLIPNeRF}. The CLIP model is replaced with Alpha-CLIP, and tests are conducted with and without background augmentation.
    }
    \label{fig:region_3d_object_generation}
    \vspace{-5mm}
\end{figure}

% \noindent \textbf{Alpha-CLIP in Point-E.} Point-E~\cite{Point-E} can achieve image-to-3D through conditioning diffusion model on CLIP vision feature of a single image. We use its base-40M model that takes whole image feature as condition and replace the original image encoder CLIP ViT-L/14 with Alpha-CLIP. We demonstrate that Alpha-CLIP does help in two cases: 1) Sometimes Point-E generates point cloud with some part missing. User can highlight the missing part in condition image to remind diffusion model to pay more attention to that part and fix the problem. 2) User can highlight the part that need to be emphasized on 2D image to tell Point-E to spend more points on that part(1024 points in total in base model). The results are shown in \cref{fig:region_3d_object_generation} with more result in \cref{sec:more_visualization_3d}.
\noindent \textbf{Alpha-CLIP in Point-E.} Point-E~\cite{Point-E} can achieve image-to-3D through conditioning diffusion model with CLIP image feature. We replace the CLIP ViT-L/14 image encoder of the Point-E base-40M model with our Alpha-CLIP. We demonstrate that Alpha-CLIP is helpful in two cases: 1) When Point-E generates the point cloud with some parts missing, users can highlight the missing part in the condition image to remind the diffusion model to pay more attention to that part and fix this missing parts problem. 2) Users can highlight the part that needs to be emphasized on the 2D image. Point-E will spend more points on the highlighted part (with 1024 points in total in the base model). The results are shown in \cref{fig:region_3d_object_generation} with more results in \cref{sec:more_visualization_3d}.

% \noindent \textbf{Alpha-CLIP in PureCLIPNeRF.} We input the rendered images with alpha channels obtained from density integration of NeRF~\cite{NeRF} into Alpha-CLIP. When optimizing the object with Alpha-CLIP, gradient can flow back from alpha channel to help generating better result. As shown in \cref{fig:region_3d_object_generation}, we find that PureCLIPNeRF generates objects that closely align with the provided textual prompts(especially bolded text) in terms of shape and color when replacing CLIP with Alpha-CLIP. Furthermore, there is an enhancement in the overall coherence of the generated objects, coupled with notable aesthetic qualities. We attribute this phenomenon to Alpha-CLIP's enhanced capability in optimizing density parameters of 3D representations directly and focus only on foreground area, which helps to generate object that is more coherent and closely match the input text.
\noindent \textbf{Alpha-CLIP in PureCLIPNeRF.} We input the rendered images with alpha channels obtained from density integration of NeRF~\cite{NeRF} into Alpha-CLIP. When optimizing the object with Alpha-CLIP, the gradient can flow back from the alpha channel to help generate better results. As shown in \cref{fig:region_3d_object_generation}, we find that PureCLIPNeRF generates objects that closely align with the provided textual prompts(especially bolded text) in terms of shape and color when replacing CLIP with Alpha-CLIP. Furthermore, there is an enhancement in the overall coherence of the generated objects, coupled with notable aesthetic qualities. We attribute this phenomenon to Alpha-CLIP's enhanced capability in optimizing density parameters of 3D representations directly and focusing only on the foreground area, which helps to generate an object that is more coherent and closely matches the input text. 

Background augmentation in PureCLIPNeRF~\cite{PureCLIPNeRF} inherited from Dream Fields~\cite{dream_fields} is a vital step to improve the consistency of objects, making them less diffuse compared to the first column in \cref{fig:region_3d_object_generation}. However, this process is time-consuming as each augmented image has to go through CLIP to get optimization direction. We thus test the capabilities of Alpha-CLIP without background augmentations. Results are presented in the second column of~\cref{fig:region_3d_object_generation}. We observe that in most cases, using Alpha-CLIP without background augmentation produces objects that are clearer and better aligned with the given text than the original CLIP with 2x faster speed. Quantitative results and More visualizations are in \cref{sec:more_visualization_3d,sec:app_3d_qua} \vspace{-1mm}

% \subsection{Failure Cases}

% \begin{figure}[t]
%     \centering
%     \includegraphics[width=\linewidth]{figure/failed_case.pdf}
%     \caption{\textbf{Alpha-CLIP failed when object is too small} When prompted region is too small, Alpha-CLIP failed to focus on them and may only output global information of the whole image.}
%     \label{fig:failed_case}
% \end{figure}

% Due to the limited resolution of the CLIP model itself, we also observed failure cases during testing where small objects are overlooked. Some cases are illustrated in \cref{fig:failed_case}. When the region of interest is too small, the model fails to focus on the specified area and may instead extracts information globally from the entire image.
% Table generated by Excel2LaTeX from sheet 'Sheet1'

% {
% \setlength{\tabcolsep}{5.4mm}
% \begin{table}[t]
%   \centering
%   \caption{\textbf{Zero Shot classification on ImageNet-S with different model scales} trained on GRIT-1m}
%   \vspace{-2mm}
%   \scalebox{0.9}{
%     \begin{tabular}{ccc}
%     model & Original CLIP & Alpha-CLIP \\
%     \hline
%     ViT-B/16 & 66.48 & 68.31\textbf{(+1.83)} \\
%     ViT-L/14 & 73.48 & 77.22\textbf{(+3.74)}	 \\
%     ViT-L/14@336 & 74.29 & 78.15\textbf{(+3.86)}	 \\
%     \hline
%     \end{tabular}%
%     }
%   \label{tab:scalability}%
% \end{table}%
% }

\section{Limitation and Future Direction}
\vspace{-1.5mm}
% \noindent While Alpha-CLIP demonstrates effective performance in various scenarios requiring region focus, its current structure and training process limit its capability to focus on multiple objects or model relationships between different objects. Furthermore, the current training methodology restricts the alpha channel from generalizing beyond intermediate values, apart from the binary values of 0 and 1. As a result, users are unable to specify the strength of attention. We plan to address these limitations in future work and expand the model resolution. enhancing the capabilities of Alpha-CLIP and extending its applicability to a broader range of downstream tasks.

While Alpha-CLIP demonstrates effective performance in various scenarios requiring region focus, its current structure and training process limit its capability to focus on multiple objects or model relationships between different objects. Furthermore, the current training methodology restricts the alpha channel from generalizing beyond intermediate values, apart from the binary values of 0 and 1. As a result, users are unable to specify the amplitude of attention. Another limitation both lays in our Alpha-CLIP and original CLIP is low resolution, which hinder the way for Alpha-CLIP to recognize small object. We plan to address these limitations in future work and expand the CLIP input resolution. We believe these future directions are pathways to augment Alpha-CLIP's abilities and broaden its utility across diverse downstream tasks. \vspace{-1.5mm}

\section{Conclusion}

% {\bf Limitations}.\shu{discuss limitations}

% {\bf Societal Impacts}.\shu{discuss societal impacts}
\vspace{-1.5mm}
In this work, We propose the Alpha-CLIP model, which introduces an additional alpha channel to specify the regions of interest. Trained on millions of RGBA region-text pairs, Alpha-CLIP not only exhibits excellent region-focus capabilities but also ensures its output space remains consistent with the original CLIP model. This consistency allows seamless replacement in various downstream applications of CLIP. We demonstrate that when prompted with specific regions of interest, Alpha-CLIP shows improved zero-shot recognition abilities and verifies its usefulness in many downstream tasks. The applications of CLIP extend far beyond the scope of this article. We hope that Alpha-CLIP will be applicable in more scenarios when foreground regions or masks are available.

% During the training process, we observe the phenomenon as shown in Table \ref{tab:scalability}: under the same training conditions, larger models achieve greater improvement in classification scores. We attribute this occurrence to an increase in resolution. Existing visual encoders tend to gradually increase feature resolution (such as the SAM model using 64$\times$64 attention) Therefore, we hope that in the future, as we improve the resolution of the CLIP-Vision Encoder and also scale up the amount of data for RGBA-region text pairs to obtaining a larger and better Alpha-CLIP Visual Encoder.

% \clearpage
% \newpage

% \setcounter{page}{1}

\appendix

% % {
% %        \twocolumn[
% %         \centering
% %         \Large
% %         \textbf{\thetitle}\\
% %         \vspace{0.5em}{\bf Supplementary Material} \\
% %         \vspace{1.0em}
% %        ] %< twocolumn
% % }

% \clearpage
% % \setcounter{page}{1}
% \maketitlesupplementary

% \section*{Appendix}

\begin{table}[b]
\begin{minipage}{\textwidth}
  \centering
    \begin{tabular}{p{7.72em}|ccccccccccc}
    \hline
    \multicolumn{1}{c}{sample ratio $r_s$} & 0.0     & \textbf{0.1} & 0.2   & 0.3   & 0.4   & 0.5   & 0.6   & 0.7   & 0.8   & 0.9   & 1.0 \\
    \multicolumn{1}{c}{top1-Acc}  & 68.06 & \textbf{68.25} & 67.87 & 67.71 & 67.83 & 67.74 & 67.37 & 66.87 & 66.39 & 64.94 & 63.96 \\
    \hline
    \end{tabular}
  \caption{\textbf{Sample ratio search experiment.} We search sample ratio with a step of 0.1. Test metric is zero-shot classification top1 accuracy on ImageNet-S~\cite{gao2022large}. As we find $r_s=0.1$ produce best result.}
  \label{tab:sup_sample_ratio}
\end{minipage}
\end{table}

\begin{table}[htbp]
\begin{minipage}{\textwidth}
  \centering
    \begin{tabular}{ccccccccc}
    \hline
    unfreeze block nums & 0     & 2     & 4     & 6     & 8     & 10    & \textbf{12} &  full-tuning on ori CLIP \\
    top1-Acc & 63.61 & 64.73 & 65.63 & 66.59 & 67.09 & 68.07 & \textbf{68.27} & 66.52(+0.04) \\
    \hline
    \end{tabular}%
    \caption{\textbf{Number of unfreeze block search experiment.} We search number of learnable Transformer block number. Test metric is zero-shot classification top1 accuracy on ImageNet-S~\cite{gao2022large}. As we find that unfreeze the whole CLIP image encoder generate the best result.}
    \label{tab:unfreeze_layer_num}%
\end{minipage}
\end{table}%

\section{Training Detail}

\subsection{Hyperparameter}
\label{app_hyperparameters}
\noindent \textbf{Basic hyperparameters.} 
The training process utilizes a batch-size of 4096 for all scales of CLIP models. We use 8 A100-80G GPUs for ViT-B/16, 64 GPUs for ViT-L/14, and 128 GPUs for ViT-L/14@336px. The training process utilizes mixed-precision float16 for acceleration. The temperature coefficient $\tau$ for CLIP is fixed to the value obtained after the completion of the original CLIP training. The optimizer chosen is AdamW~\cite{adamw} with a weight decay of 2e-2. Regarding learning rates, the learning rate for the convolutional kernels accepting alpha channel input is set to 2e-4, while the rest of the layers have a learning rate of 2e-6, employing a cosine learning rate scheduler. For GRIT-1m, the training lasts 6-8 epochs, whereas GRIT-20m is trained for 2 epochs.

\noindent \textbf{Whole image sample ratio.} 
Due to our desire to preserve the original CLIP's recognition ability for the entire image, in the training of Alpha-CLIP on GRIT~\cite{Kosmos-2}, we sample a portion of RGBA-text pairs and set the alpha channel to all 1(indicating region over the entire image). The text is replaced with the original full-image caption. We use the ViT-B/16 model and train on GRIT for 4 epochs, varying the sampling ratio. Zero-shot classification on Imagenet-S is used as the evaluation metric. The experimental results, as shown in \cref{tab:sup_sample_ratio}, indicate that training without sampling a proportion of the entire image-text pair performs worse than choosing a small number of images that require full-image attention. However, excessively emphasizing full-image attention during training significantly impairs the model's capability. Based on these results, a sample ratio of 0.1 is chosen for the experiments in the main text.

\begin{table}[htbp]
  \centering
    \begin{tabular}{ccccc}
    \toprule
          & test with gt mask & all\_0 & all\_1 & ori\_clip \\
    \midrule
    whole\_1 & 77.41 & 72.53 & 73.37 & \multirow{2}[0]{*}{73.48} \\
    whole\_0 & 74.99 & 73.45 & 73.27 &  \\
    \bottomrule
    \end{tabular}%
    \caption{\textbf{Different strategy for whole image perception experiment.} Test metric is zero-shot classification top1 accuracy on ImageNet-S~\cite{gao2022large}}
    \label{cross_test_whole_image_setting}%
\end{table}%

\noindent \textbf{Whole image perception setting.} 
In accordance with the definition of transparency in the 2D image, setting alpha to all 1 is indicative of the requirement for CLIP to focus on the entire image. However, due to the absence of bias in the first layer convolution of CLIP, which consists only of weights, an input with an alpha channel of all 0 maintains CLIP's original state, contrary to the definition of image transparency. To determine an optimal approach, we conducted training and testing under different configurations. During training, we utilized image-text pairs with the entire image set to all 0 and all 1, each with a sample ratio of 0.1. During testing, we cross-validated the classification accuracy with alpha channels set to all 0 and all 1. Experiment results are presented in \cref{cross_test_whole_image_setting}. Our observations indicate that configuring the training and inference with the alpha channel set to all 1 achieves the best perception performance. Therefore, we consistently adopt this configuration in the main section of the paper, utilizing an all-1 alpha input when Alpha-CLIP is required to focus on the entire image.

\noindent \textbf{Unfreeze block number.}
We search through the number of layers to unfreeze when training Alpha-CLIP on RGBA-text pairs that create the best result. we test on ViT-B/16(with 12 attention blocks) and train on GRIT-1m~\cite{Kosmos-2} for 4 epochs. We select the zero-shot classification top1 accuracy on ImageNet-S~\cite{gao2022large} as our test metric. The result is shown in \cref{tab:unfreeze_layer_num}. We also test full fintuning of original CLIP without alpha channel on GRIT~\cite{Kosmos-2} dataset, which only gets negligible improvement, proving that the improvement contributes to the input focus region through alpha channel instead of training data.
\newpage
 We unfreeze the number of transformer blocks from 0 to 12 (full) with a step of 2. Results show that as the number of unfrozen blocks increases, the classification accuracy increases steadily. We also test LoRA~\cite{lora}, but it does not work well compared with full model fintuning. So we consider full model tuning in the main part of the paper.

\subsection{Ablation on Data Volume}
\label{app_ablation_data_volume}
\begin{figure}[]
    \centering
    \includegraphics[width=\linewidth]{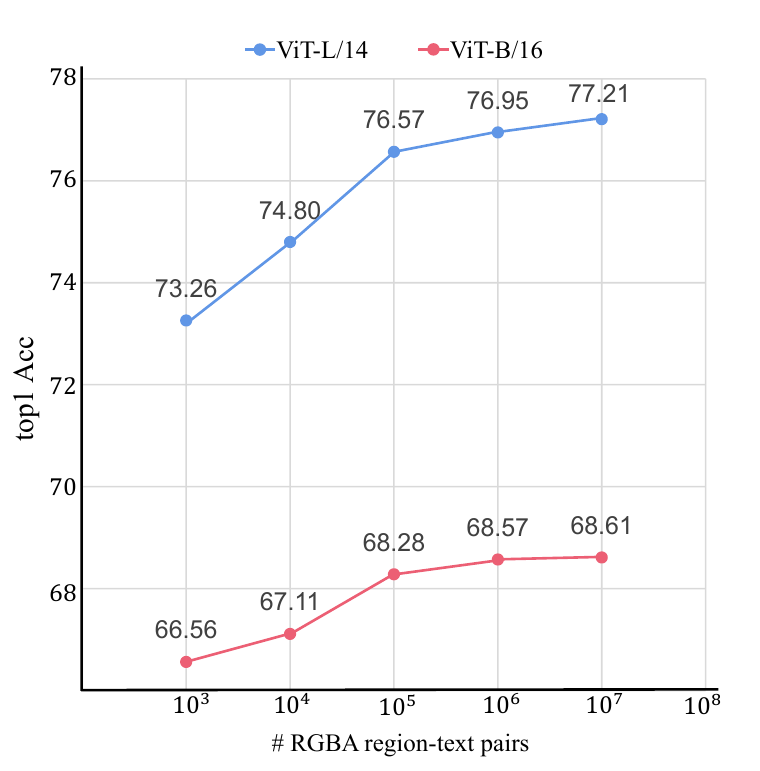}
    \caption{\textbf{Zero-shot ImageNet-S~\cite{gao2022large} classification accuracy w.r.t training data volume.} two different scale ViT model of CLIP~\cite{CLIP} are tested.
    }
    \vspace{-3mm}
    \label{fig:data_ablation}
\end{figure}

We examine the efficacy of data volume in enhancing the training of robust models through an ablation study. Our ablation involved the training of ViT-B/16 and ViT-L/14 using RGBA region-text pairs, with data quantities ranging from 1k to 10M. We use the zero-shot top-1 accuracy on ImageNet-S as our evaluation criterion. As illustrated in \cref{fig:data_ablation}, increasing data volume corresponds to a concurrent improvement in the model's classification accuracy. Notably, larger ViT models exhibited a more substantial performance boost in comparison to their smaller counterparts throughout this process.

\section{Different implementation of MaskCLIP}
\label{app_MaskCLIP}
\noindent To the best of our knowledge, two methods propose to use the attention mask to guide the CLIP visual model to pay more attention to the foreground area. We test these two method respectively. Because these two methods can only do masking at the feature level as [H, W]
([14, 14] for ViT-B/16, [16, 16] for ViT-L/14), we first do max pooling on the binary mask $M$ to make it match the size of the feature map.

\begin{equation}
\begin{split}
m = \text{MaxPooling}(M)
\end{split}
\label{eq:clip_loss}
\end{equation}

\noindent \textbf{Mask Area guided last attention.} As~\cite{MaskCLIP} proposes using 1$\times$1 conv layer to get feature space level 2D semantic classification map, we use the same idea of the attention mask to set \texttt{[cls]} token only to calculate attention with foreground area patches. In other words, we use $m$ to guide the last attention calculation. We report this result in the main part of this paper as it shows better results than the original CLIP.

\begin{table}[t]
  \centering
    \begin{tabular}{ccc|cc}
    \multicolumn{3}{c}{model} & ViT-B/16 & ViT-L/14 \\
    \hline
    \multicolumn{2}{c}{\multirow{2}[0]{*}{Original CLIP}} & top1  & 66.48  & 73.48  \\
    \multicolumn{2}{c}{} & top5  & 88.90  & 91.60  \\
    \hline
\multicolumn{2}{c}{\multirow{2}[0]{*}{MaskCLIP~\cite{attention_mask}}} & top1  & 53.61  & 65.28  \\
    \multicolumn{2}{c}{} & top5  & 76.47  & 87.25  \\
    \hline
\multicolumn{2}{c}{\multirow{2}[0]{*}{Alpha-CLIP}} & top1  & 68.89  & 77.41  \\
    \multicolumn{2}{c}{} & top5  & 90.51  & 94.45  \\
    \hline
    \end{tabular}%
    \caption{\textbf{Zero-shot classification on ImageNet-S\cite{gao2022large}.} Comparison using method proposed in~\cite{attention_mask}.}
    \label{tab:sup_maskclip}%
\end{table}%

\noindent \textbf{Relative Mask Attention.} This method is proposed in~\cite{attention_mask}, and is used in~\cite{ODISE}, which introduces "Mask Patch Tokens" to do the same attention as \texttt{[CLS]} token but only attach to those patches that contain foreground area. We test this method but it does not produce good results on ImageNet-S~\cite{gao2022large} as shown in \cref{tab:sup_maskclip} as it is used in the segmentation task to classify each semantic mask(area) of a whole image, but ImageNet-S~\cite{gao2022large} only cares about a single prominent foreground object in most cases. So we do not report this result in the main part of this paper. 

\section{Effectiveness of Classification Data}
\label{app_ablation_mixture_of_data}
\noindent While Grounding data holds more promising prospects in the future, especially with the advent of more powerful grounding and segmentation models, we demonstrate that, at the current stage, leveraging large-scale manually annotated classification datasets like ImageNet~\cite{Imagenet21k} and constructing RGBA region-text pairs using the pipeline shown in \cref{fig:overall} still significantly benefits Alpha-CLIP in achieving enhanced Region-Perception capabilities.

\subsection{Zero-shot Classification on COCO} 
\begin{table}[htbp]
  \centering
  \scalebox{0.90}{
    \begin{tabular}{cc|cc}
    \multicolumn{2}{c}{model} & ViT-B/16 & ViT-L/14 \\
    \hline
    \multirow{3}[0]{*}{masking} & CLIP  & 49.42  & 54.43  \\
          & Alpha-CLIP$_g$ & 49.27 & 56.45 \\
          & Alpha-CLIP$_{g+c}$ & \textbf{53.39} & \textbf{58.84} \\
    \hline
    \multirow{3}[0]{*}{no masking} & CLIP  & 64.21  & 67.65  \\
          & Alpha-CLIP$_g$ & 61.57 & 67.44 \\
          & Alpha-CLIP$_{g+c}$ & \textbf{71.08} & \textbf{77.56} \\
    \hline
    \multirow{3}[0]{*}{ImageNet-S top1} & CLIP  & 66.48  & 71.48  \\
          & Alpha-CLIP$_g$ & 68.89 & 77.41 \\
          & Alpha-CLIP$_{g+c}$ & \textbf{69.40} & \textbf{77.80} \\
    \hline
    \end{tabular}%
    }
    \caption{\textbf{Zero-shot classification results on COCO.} Our Alpha-CLIP also achieve significant improvement on zero-shot Instance-COCO~\cite{COCO} classification tasks.}
    \label{tab:coco_table}%
\end{table}%
\noindent In addition to natural image classification tests, there are scenarios where there is a need to crop or mask objects in images\cite{OvarNet} \cite{ReCLIP} \cite{MaskAdaptedCLIP}. Therefore, we conducted classification tests for Alpha-CLIP in such scenarios using the validation set of the Instance-COCO~\cite{COCO} dataset, which consists of 80 classes. We cropped objects using ground-truth bounding boxes and enlarged them by 1.5 times (referred to as COCO crop). We conduct tests in two scenarios: masking (setting the background to a solid color) and no masking (using the original background). To prevent results from being dominated by the most frequent classes, we use the mean of per-class accuracy as the evaluation metric. To ensure that Alpha-CLIP is adapted to images with backgrounds replaced by solid colors, we incorporate object-centric image data (from the lower branch of \cref{fig:overall} into the training data for this scenario. This data is generated from the top 460k RGBA-region text pairs auto-generated from ImageNet-21k~\cite{Imagenet21k}, and we include it in the pairs generated from GRIT-1M~\cite{Kosmos-2} as the training dataset for Alpha-CLIP. Results are shown in \cref{tab:coco_table}. We compare it with the baseline method trained on GRIT-1m only and find a huge improvement for cropped image classification. We also test its classification accuracy on ImageNet-S~\cite{gao2022large}, and the result even surpasses models trained on GRIT-20m. We contribute this to the human annotation of fine-grained class labels of ImageNet~\cite{Imagenet21k} dataset.

\subsection{Different version Alpha-CLIP in REC}
\begin{table}
\centering
    % \vspace{-4mm}
% \vspace{-6mm}
\scalebox{0.70}{
\begin{tabular}{cccc|ccc|cc}
    \toprule
\multirow{2}{*}{\textbf{Data}} &\multicolumn{3}{c|}{\textbf{RefCOCO}} & \multicolumn{3}{c|}{\textbf{RefCOCO+}} & \multicolumn{2}{c}{\textbf{RefCOCOg}} \\
 & Val & TestA & TestB & Val & TestA & TestB & Val & Test \\
\midrule
GRIT-1M & \textbf{56.1} & \textbf{63.4} & 48.9 & 55.1 & 62.6 & 45.1 & 60.3 & 60.6 \\
GRIT-1M + IN & 55.7 & 61.1 & \textbf{50.3} & \textbf{55.6} & \textbf{62.7} & \textbf{46.4} & \textbf{61.2} & \textbf{62.0} \\
    \bottomrule \\
\end{tabular}}

% \vspace{-2em}
%
\caption{\textbf{Zero-shot REC results of Alpha-CLIP with different pretraining data.} We compare the results of using only grounding data with adding classification data. We report top-1 accuracy (\%).}
\label{tab:append_ref_exp}
\end{table}

\noindent To investigate the effectiveness of the classification data on REC, we conduct experiments comparing Alpha-CLIP pretrained solely on the grounding data with a combination of classification and grounding data. We use an ensemble of ViT-B/16 and ViT-L/14 backbones, with grounding data sourced from GRIT-1M~\cite{Kosmos-2} and classification data from ImageNet-21k~\cite{Imagenet21k}. As shown in \cref{tab:append_ref_exp}, on the majority of benchmarks, using classification data yields better results compared to models that are not pretrained with it.

\section{Other CLIP masking baselines}
\label{sec:app_masking_baseline}
\begin{figure}[]
    \centering
    \includegraphics[width=\linewidth]{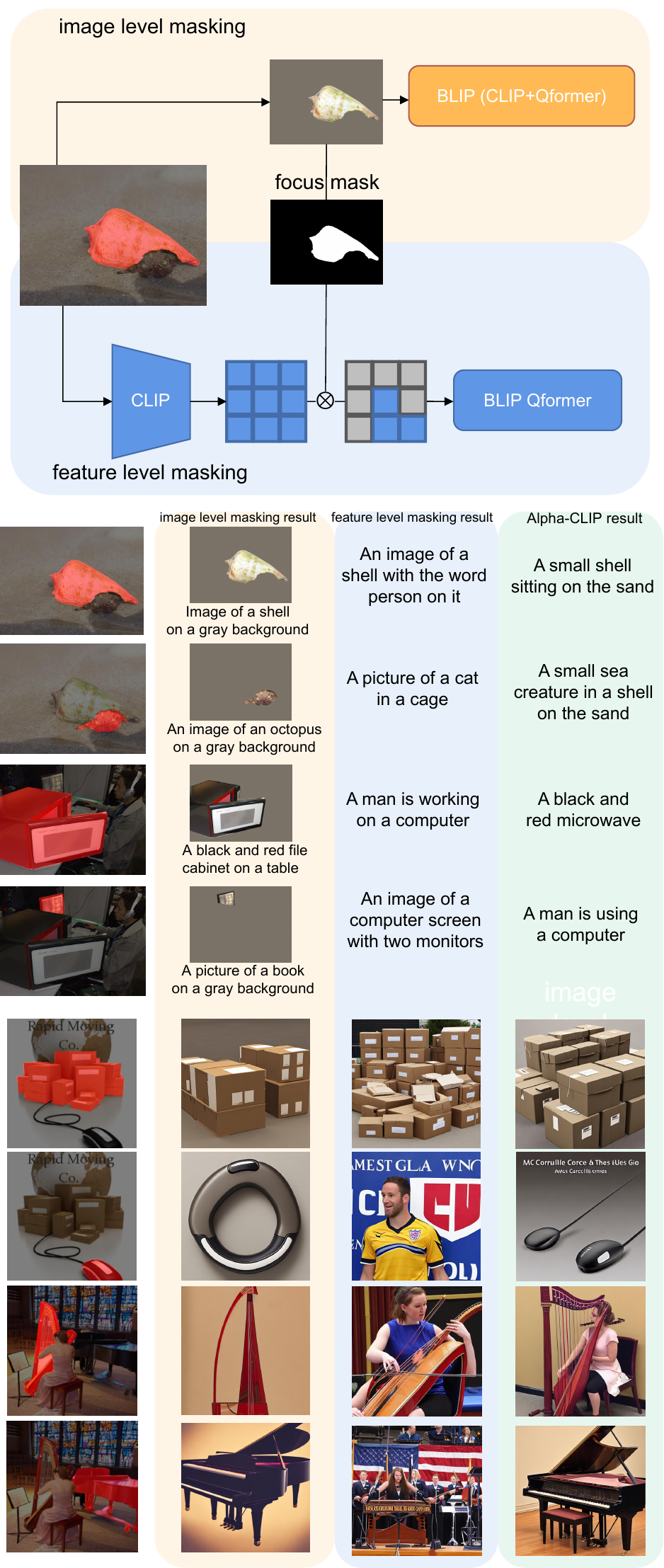}
    \caption{\textbf{Two baselines of image level masking and feature level masking and their comparison with Alpha-CLIP.} It is worth noticing that we use BLIP-2~\cite{BLIP-2} structure with Q-former as presented in the baseline pipeline. Alpha-CLIP and these two masking approaches can also adapt to structures that only have the projection layer, like LLaVA~\cite{LLaVA} and miniGPT-v2~\cite{miniGPT-v2}
    }
    \label{fig:mask_ablation}
\end{figure}
\noindent There are simple ways to make CLIP focus on user-specified regions without modifying the weights of the CLIP model. We test two possible approaches here. Namely Image-level masking and feature-level masking. We test on these simple baselines and make comparisons with Alpha-CLIP. As shown in \cref{fig:mask_ablation}. It is worth noticing that we only draw structure using the Q-former proposed by BLIP-2. But they can also be adapted to other VL systems that use CLIP image encoder as the visual backbone like LLaVA~\cite{LLaVA, llava-1.5} and miniGPT4-v2~\cite{miniGPT-v2}.

\noindent \textbf{Image Level Masking} means simply masking the background region by directly setting these regions to pure color. We choose the color same as MaskAdaptedCLIP~\cite{MaskAdaptedCLIP}.

\begin{figure}[]
    \centering
    \includegraphics[width=\linewidth]{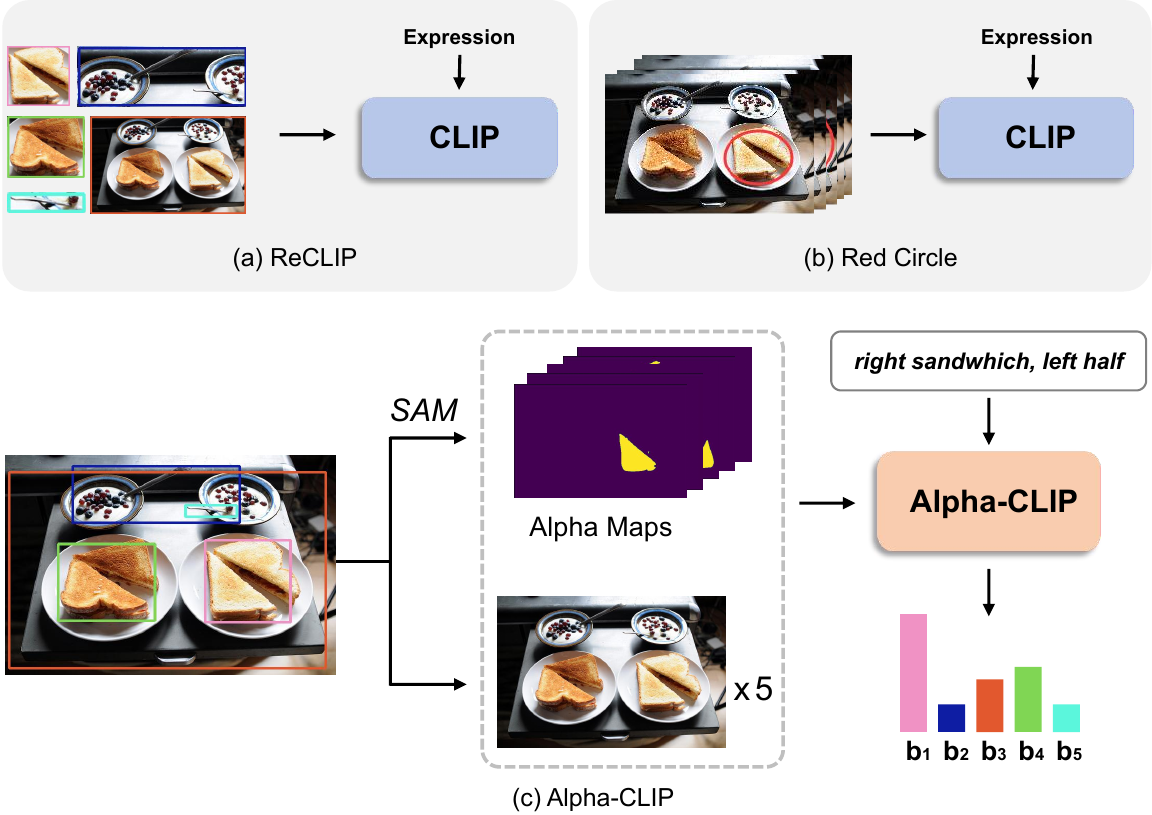}
    \caption{\textbf{Model pipeline of Alpha-CLIP in Zero-shot REC Tasks.}
    We compared our model (as illustrated in a detailed flowchart in the lower part) with two other baselines~\cite{ReCLIP, circleCLIP} (represented by a concise flowchart in the upper part).
    }
    \label{fig:rec_pipeline}
\end{figure}

\noindent \textbf{Feature Level Masking} is another method that applies masking on the feature level. As shown in \cref{fig:mask_ablation}, we use max pooling to downsample the image level mask to feature level coarse-grained mask, and use this mask to do element-wise product to set the features that belong to the background to zero. This method has been proven useful in Ellite~\cite{wei2023elite} when the object is in the center of the image and occupies a large space.

Results of the two masking methods are shown at the top of \cref{fig:mask_ablation}. We use the same settings as in the main section of the paper. BLIP-2~\cite{BLIP-2}, CLIP-L/14+flant5xl is used for captioning, BLIP-Diffusion~\cite{BLIP-Diffusion}, CLIP-L/14+stable-diffusion~\cite{Stable-Diffusion} is used for Image generation. The first column represents the original image and the area that needs to be focused, the second column represents the results of using image-level masking, the third column represents the results of using feature-level masking, and the fourth column represents the results of our Alpha-CLIP. The first four lines are image captioning results, and the last four lines are image variation results. As can be seen from \cref{fig:mask_ablation}, image-level masking will lose the context information of the object, causing its semantics to be incorrect or blurred and cannot produce good results; while feature-level masking can sometimes produce better results, but rough masking directly on the feature level may cause unpredictable behaviors, such as generating pictures with completely irrelevant semantic information, or being dominated by the semantics of the main objects in the picture. In contrast, Alpha-CLIP can produce better results because it is pre-trained on millions of RGBA-text pairs. These two feature masking methods also destroy the features of other areas of the image to a greater extent and completely lose information about the other part of the image, and therefore fail in simple reasoning problems that need to involve the relationship between objects and the environment. In the meantime, Alpha-CLIP can highlight the region that needs to be focused on with features of the remaining areas in the image better preserved.

\section{Zero-shot REC with Alpha-CLIP Implementation Details}
\label{sec:app_zero_shot_REC_imp}
\label{sec:REC pipeline}

\begin{table}
% \vspace{-4mm}
\scalebox{0.75}{
\begin{tabular}{lccc|ccc|cc}
    \toprule
\multirow{2}{*}{\textbf{PP}} & \multicolumn{3}{c|}{\textbf{RefCOCO}} & \multicolumn{3}{c|}{\textbf{RefCOCO+}} & \multicolumn{2}{c}{\textbf{RefCOCOg}} \\
& Val & TestA & TestB & Val & TestA & TestB & Val & Test \\
\midrule 
 B & 56.8 & \textbf{63.7} & 49.4 & 56.2 & 63.6 & 45.9 & 59.9 & 61.7 \\
 B \textbar~C & \textbf{57.0} & 62.8 & 50.6 & \textbf{57.1} & 63.7 & 48.0 & \textbf{64.0} & 64.1 \\
 B \textbar~C \textbar~G & \textbf{57.0} & 63.0 & \textbf{51.0} & 56.9 & \textbf{64.0} & \textbf{48.5} & 63.6 & \textbf{64.3} \\
    \bottomrule \\
\end{tabular}
}
\vspace{-2mm}
\caption{\textbf{Zero-shot REC results of Alpha-CLIP with different image preprocessing methods.} We make comparisons across RefCOCO, RefCOCO+, and RefCOCOg datasets. \textbf{PP}: Preprocessing methods. ``B'' denotes original input and blurring operation. ``C'' denotes cropping. ``G'' denotes grayscaling.}
\vspace{4mm}
% \vspace{-2em}
%
\label{tab:ref-exp-append}
\end{table}

\begin{figure}[]
    \centering
    \includegraphics[width=\linewidth]{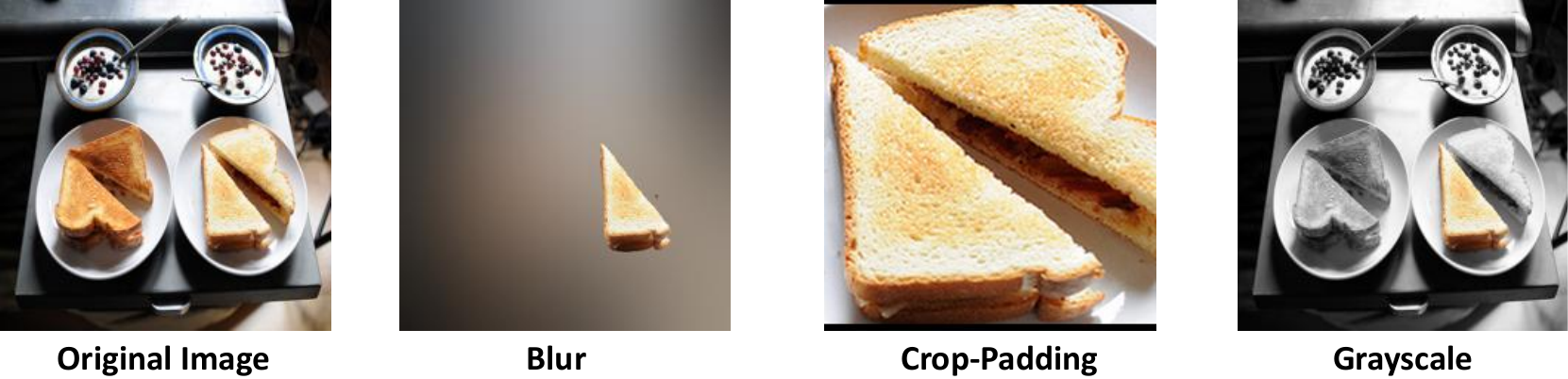}
    \caption{\textbf{Visualization of different image preprocessing operations.} 
    In our basic approach, we only utilize the original image and blurring. Additionally, we supplement the process with cropping and grayscaling operations.
    }
    \label{fig:rec_aug}
\end{figure}

% \begin{table}
% % \vspace{-4mm}
% \scalebox{0.65}{
% \begin{tabular}{lcccc|ccc|cc}
%     \toprule
% \multirow{2}{*}{\textbf{Data}} & \multirow{2}{*}{\textbf{PP}} & \multicolumn{3}{c|}{\textbf{RefCOCO}} & \multicolumn{3}{c|}{\textbf{RefCOCO+}} & \multicolumn{2}{c}{\textbf{RefCOCOg}} \\
%  & & Val & TestA & TestB & Val & TestA & TestB & Val & Test \\
% \midrule
% \multirow{2}[0]{*}{GRIT-1M+IN} & B & 55.7 & 61.1 & 50.3 & 55.6 & 62.7 & 46.4 & 61.2 & 62.0 \\
%  & B \textbar~C & 56.1 & 60.6 & 50.3 & 56.3 & 62.7 & \textbf{48.0} & \textbf{64.1} & \textbf{64.1} \\
% \midrule
% \multirow{2}[0]{*}{GRIT-20M} & B & 56.8 & \textbf{63.7} & 49.4 & 56.2 & 63.6 & 45.9 & 59.9 & 61.7 \\
%  & B \textbar~C & \textbf{57.0} & 62.8 & \textbf{50.6} & \textbf{57.1} & \textbf{63.7} & \textbf{48.0} & 64.0 & \textbf{64.1} \\
%     \bottomrule \\
% \end{tabular}
% }
% % \vspace{-6mm}
% \caption{\textbf{Zero-shot REC results of Alpha-CLIP with different image preprocessing methods with different pretraining data.} We make comparisons across RefCOCO, RefCOCO+, and RefCOCOg datasets. \textbf{PP}: Preprocessing. ``B'' denotes original input and blurring operation. ``C'' denotes cropping method.}
% \vspace{-6mm}
% % \vspace{-2em}
% %
% \label{tab:ref-exp-append}
% \end{table}

\noindent In \cref{fig:rec_pipeline} (c), we provide a detailed illustration of the Alpha-CLIP model pipeline in zero-shot Referring Expression Comprehension. We also compare our architecture with ReCLIP~\cite{ReCLIP} and RedCircle~\cite{circleCLIP}, highlighting the differences and advantages. ReCLIP employs cropping and blurring operations to isolate image regions, which are obtained from box proposals from a detector~\cite{mattnet}. However, as shown in \cref{fig:rec_pipeline} (a), cropping results in losing relationship information between objects, leading to decreased performance. While RedCircle draws red circles on specific regions across the entire image to avoid this issue, its prompt is still coarse-grained and it alters the original image, as presented in \cref{fig:rec_pipeline} (b). We use SAM~\cite{SAM} to generate fine-grained alpha maps with the aforementioned box proposals. We input both alpha maps and original or blurred images into Alpha-CLIP and calculate similarity with referring expressions. The remaining steps closely align with ~\cite{ReCLIP}. Our basic approach ensures the complete input of images and utilizes preprocessing methods as few as possible. It is efficient and achieves excellent performance across different benchmarks, as demonstrated in \cref{tab:ref-exp}.

In addition to the preprocessing operations in our basic approach (original image and blurring), we further explore the cropping and grayscaling operations depicted in \cref{fig:rec_pipeline} (a) and (b). More specifically, for the blurring operation, our hyperparameter, namely the standard deviation ($\sigma$), is set to $\sigma=100$ ~\cite{ReCLIP}. For the cropping operation, we pad the cropping box to be a square and fill the background at the image level with zeros (i.e., black color), as shown in \cref{fig:rec_aug}. We use an ensemble of ViT-B/16 and ViT-L/14 Alpha-CLIP backbones to test, which are trained on GRIT-20M. The results, as shown in \cref{tab:ref-exp-append}, indicate the additional benefits of incorporating these operations. This underscores the strong adaptability of our model, demonstrating its ability to adapt to diverse image inputs.

% Our approach avoids both the loss of relational information caused by cropping and any modifications to the original image. It achieves excellent performance across different benchmarks, as demonstrated in \cref{tab:ref-exp}.

\begin{table}
  \centering
   \scalebox{0.95}{
  \begin{tabular}{cccc}
    \toprule
    Method & Res+Iter & R-Precision & Time \\
    \midrule
    PureCLIPNeRF $\dagger$ & 168²+10k & 85.62 & $\sim$34min \\
    $\alpha$-PureCLIPNeRF & 168²+10k & \textbf{88.89} & $\sim$36min \\
    \bottomrule
  \end{tabular}
  }
  \caption{\textbf{Quantitative Results of 3D Generation.} We compare the R-Precision of PureCLIPNeRF~\cite{PureCLIPNeRF} model using original CLIP and Alpha-CLIP, as well as the time cost to generate a single object. $\dagger$ indicates our reimplementation.}
  \label{tab:3d_quan}%
\end{table}

\section{
Quantitative Results of Nerual Field Optimization based 3D Object Generation}
\label{sec:app_3d_qua}

We evaluate the quantitative results of Alpha-CLIP in PureCLIPNeRF, using the same test method proposed in Dreamfields~\cite{dream_fields}, which includes 153 text prompts related to the COCO dataset. We measure the generated results using CLIP R-Precision. Under the same setting proposed in ~\cite{PureCLIPNeRF}, we use the Alpha-CLIP ViT-B/16 model to optimize the generated object and compare our method with the original CLIP. We test R-Precision using CLIP ViT-B/32. The results are presented in \cref{tab:3d_quan}, which shows our Alpha-CLIP can generate better objects than the original CLIP with negligible extra time consumption(test on V100-32G GPU).

\section{More Qualitative Result Visualization}

\subsection{Region-focused Image Captioning}
\label{sec:more_visualization_blip2}
\begin{figure*}[]
    \centering
    \includegraphics[width=\linewidth]{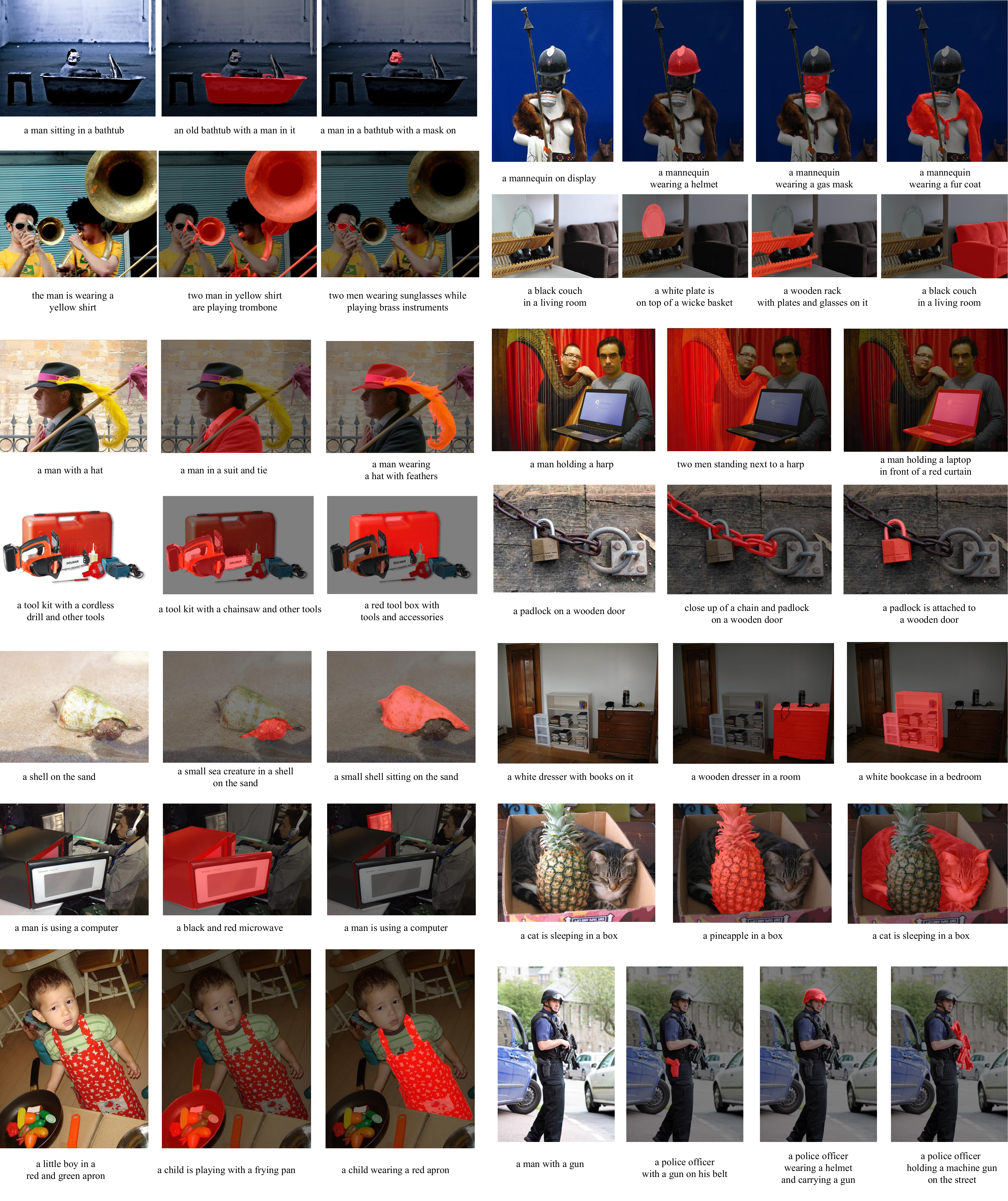}
    \caption{\textbf{More results of Alpha-CLIP used in BLIP-2~\cite{BLIP-2}.} The first row per three is the original BLIP-Diffusion generated images. Other rows represent the outcomes of Alpha-CLIP with highlighted regions marked in red. It is worth noticing that although we use ground truth mask as the highlighted region, Alpha-CLIP can actually take any region as alpha input. Users can easily get high-quality masks with the help of SAM\cite{SAM}.}
    \label{fig:append_region_image_caption}
\end{figure*}

\noindent As described in \cref{sec:alpha_clip_MLLM}, we replace original CLIP ViT-L/14 in BLIP-2~\cite{BLIP-2} with Alpha-CLIP without any post fine-tuning to generate captions. More results are shown in \cref{fig:append_region_image_caption}

\subsection{Region-focused VQA and Detailed Image Discription}
\label{sec:more_visualization_llava}
\begin{figure*}[]
    \centering
    \includegraphics[width=\linewidth]{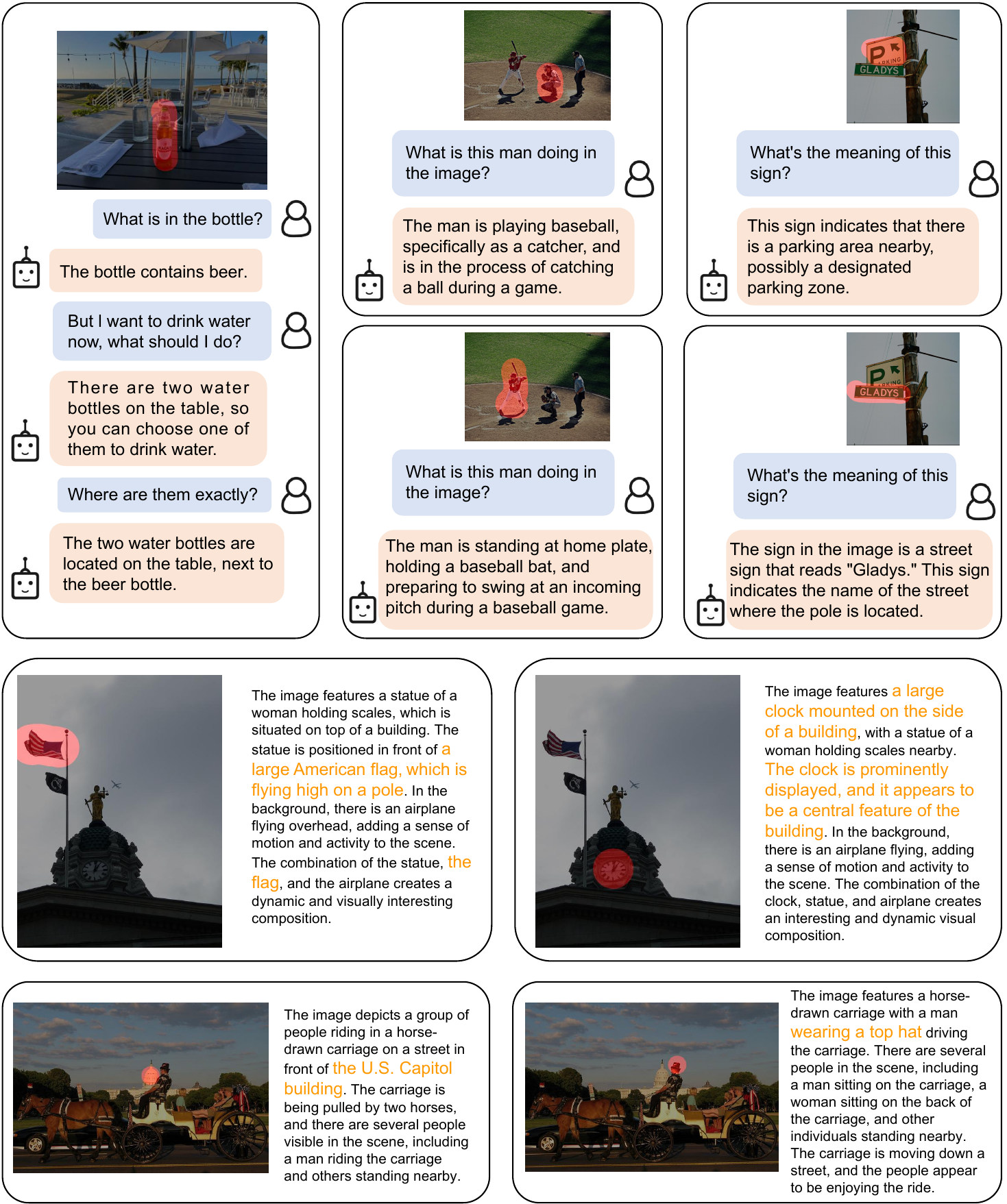}
    \vspace{-4mm}
    \caption{\textbf{More results of Alpha-CLIP used in LLaVA-1.5~\cite{llava-1.5}.} All cases shown here are made simply by replacing the original
CLIP of LLaVA-1.5~\cite{llava-1.5} with a plug-in Alpha-CLIP without further tuning. Alpha-CLIP can achieve region-based VQA and region-focused detailed image descriptions.}
    \label{fig:append_region_mllm}
\end{figure*}

\noindent As described in \cref{sec:alpha_clip_MLLM}, we replace original CLIP ViT-L/14-336px in LLaVA-1.5~\cite{llava-1.5} without any post fine-tuning. Results are shown in \cref{fig:append_region_mllm}

\subsection{Region-focused Image Variation}
\label{sec:more_visualization_blip-diff}
\begin{figure*}[]
    \centering
    \includegraphics[width=\linewidth]{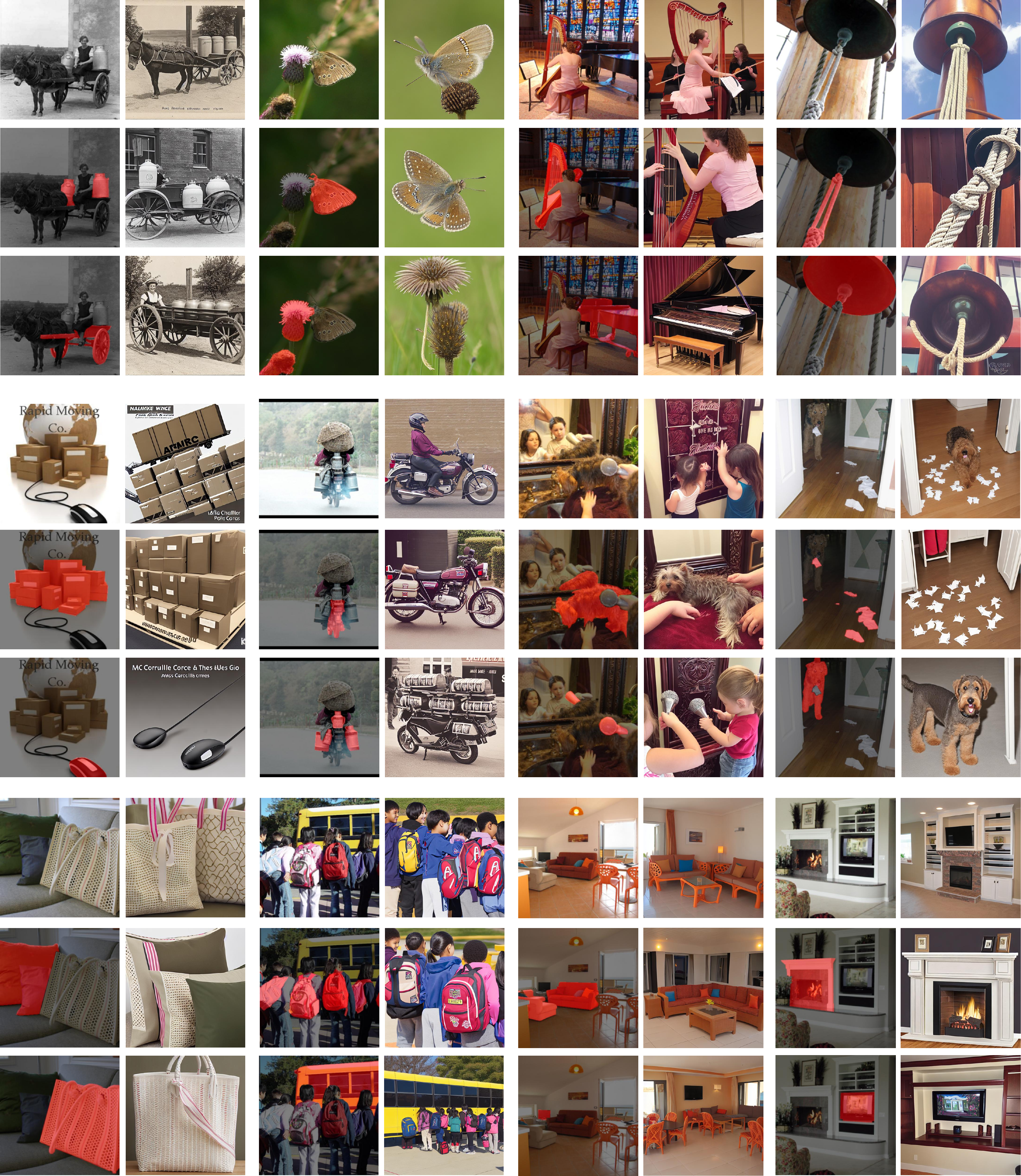}
    \caption{\textbf{More results of Alpha-CLIP used in BLIP-Diffusion~\cite{BLIP-Diffusion}.} The first row per three is the original BLIP-Diffusion generated images. Other rows represent the outcomes of Alpha-CLIP with highlighted regions marked in red. It is worth noticing that although we use ground truth mask as the highlighted region, Alpha-CLIP can actually take any region as alpha input. Users can easily get high-quality mask with the help of SAM\cite{SAM}.}
    \label{fig:append_region_image_generation}
\end{figure*}

\noindent As described in \cref{sec:alpha_clip_image_var}, we replace the original CLIP ViT-L/14 used in BLIP-Diffusion to make the condition image feature mainly focus on the user-specified area while maintaining background information. Results are shown in \cref{fig:append_region_image_generation}.

\subsection{3D Object Generation}
\label{sec:more_visualization_3d}
\begin{figure*}[]
    \centering
    \includegraphics[width=\linewidth]{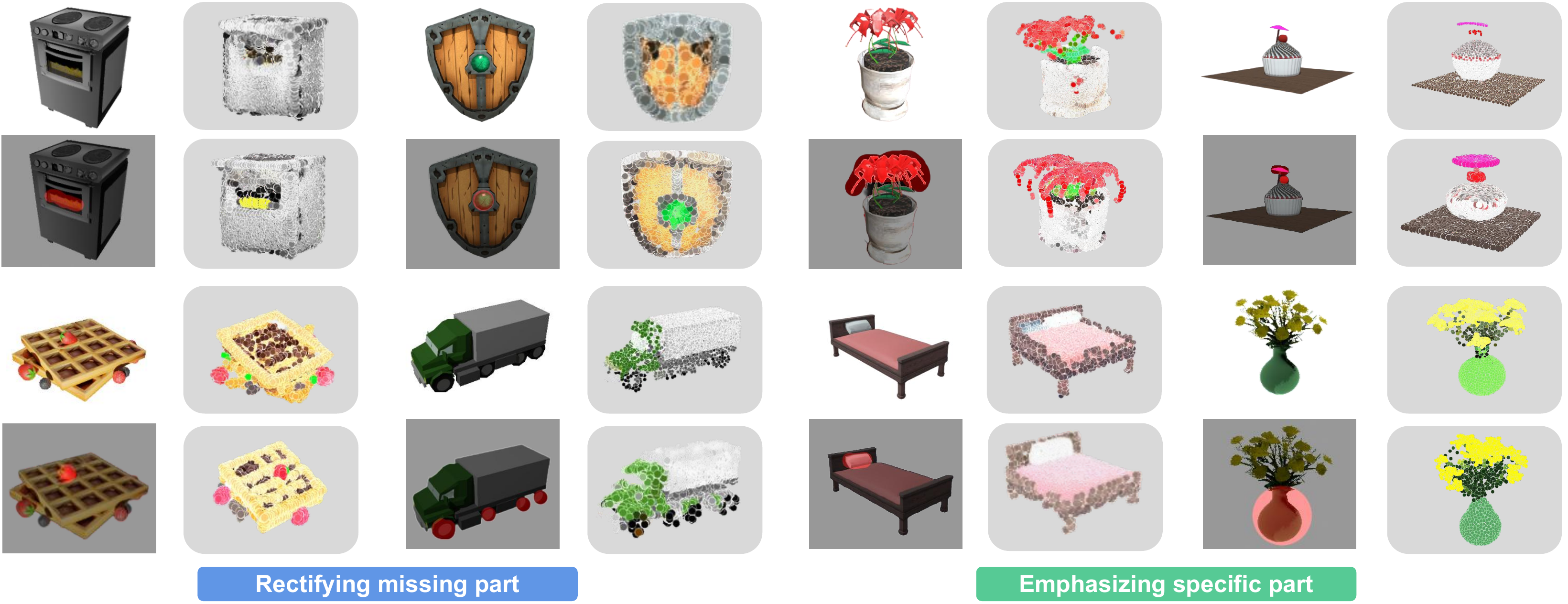}
    \caption{\textbf{More results of Alpha-CLIP used in Point-E\cite{Point-E}.} In each example, the results in the first row are 3D point clouds generated by the original CLIP, while the results in the second row are 3D point clouds generated with highlighted areas in red under the guidance of Alpha-CLIP. The left part shows Alpha-CLIP's ability to rectify missing parts, and the right part shows emphasizing specific areas using Alpha-CLIP.}
    \label{fig:3d_pointe}
\end{figure*}

\begin{figure*}[]
    \centering
    \includegraphics[width=\linewidth]{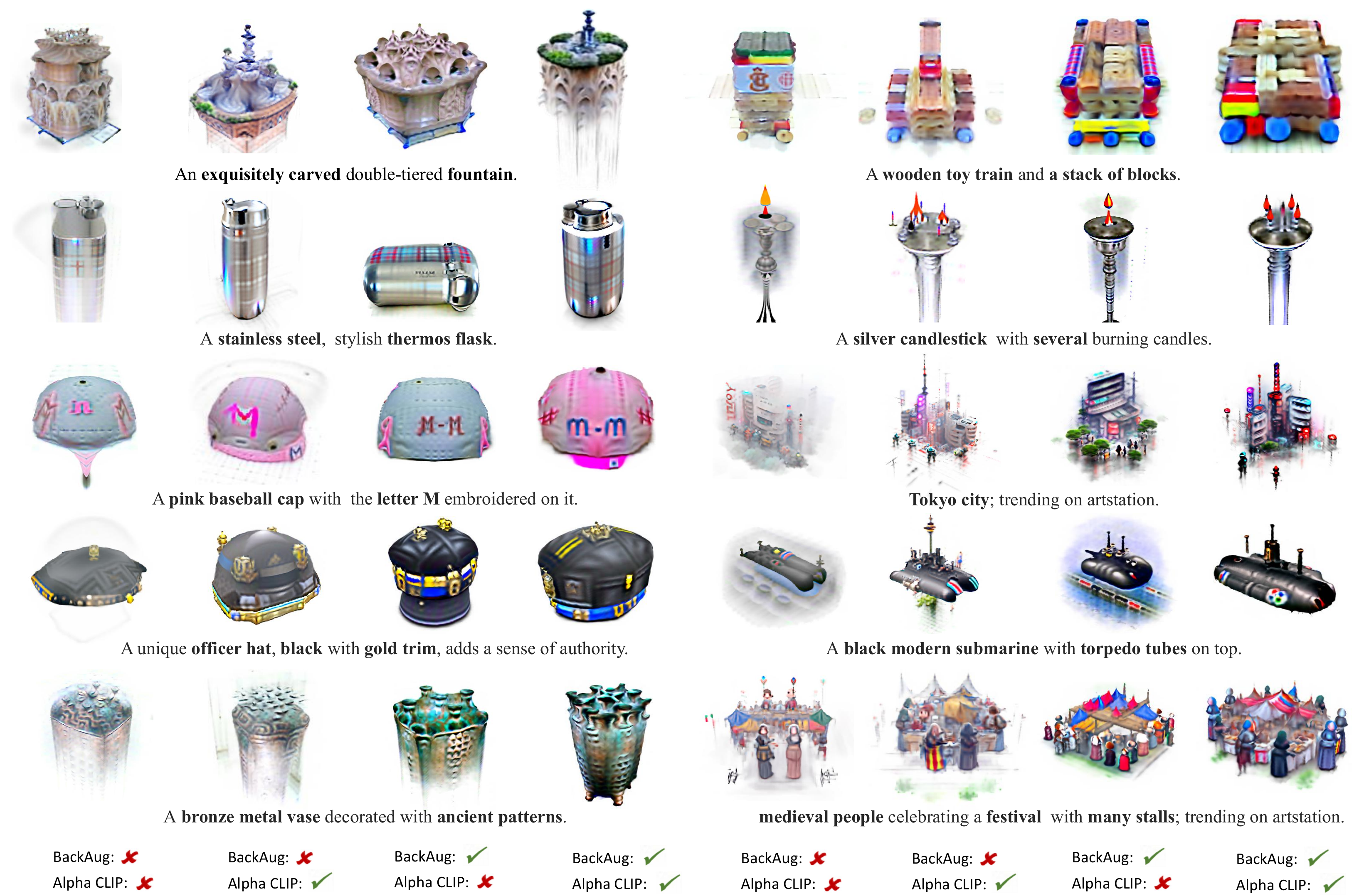}
    \caption{\textbf{More results of Alpha-CLIP used in PureCLIPNeRF\cite{PureCLIPNeRF}.} In each example, the results in the last two columns are 3D objects generated with PureCLIPNeRF under the guidance of Alpha-CLIP and the original CLIP, while the results in the first two columns are objects generated by them respectively but without background augmentations.}
    \label{fig:3d_pureclip}
\end{figure*}

\noindent Using the same setting in \cref{sec:alpha_clip_3d} Diffusion based object generation based on Point-E~\cite{Point-E} base-40M are shown in \cref{fig:3d_pointe}, where Alpha-CLIP can achieve user-defined area focus to rectify missing part or emphasizing specific part. Neural field optimization based object generation based on PureCLIPNeRF~\cite{PureCLIPNeRF} using ViT-B/16 for object optimization is shown in \cref{fig:3d_pureclip}, where Alpha-CLIP generally generates better objects than original CLIP with or without background augmentation.

\subsection{Attention map in Alpha-CLIP}
\label{sec:more_visualization_attn}
\begin{figure*}[]
    \centering
    \includegraphics[width=\linewidth]{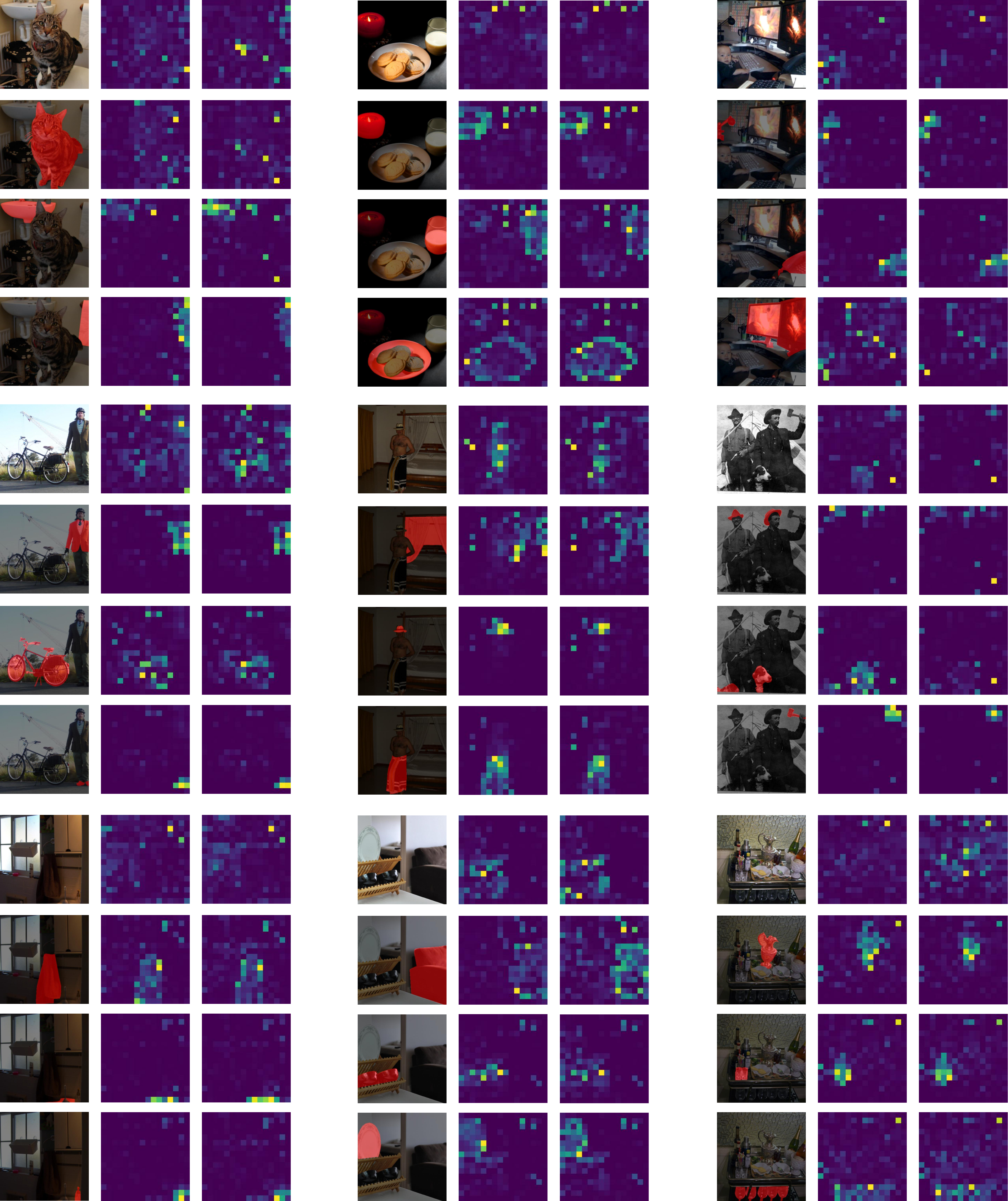}
    \caption{\textbf{Alpha-CLIP Attention map visualization.} Last transformer block attention map of \texttt{[CLS]} token with other patch tokens. each first line per four is from original CLIP~\cite{CLIP} and the other three lines are from Alpha-CLIP with user-defined focus regions marked in red. Prompting with region need focusing, Alpha-CLIP will focus on the part accordingly without compromising the original object location in feature grid.}
    \label{fig:more_attn}
\end{figure*}

\noindent Follow the spirit of DINO~\cite{dino}, we visualize Alpha-CLIP attention maps to check whether Alpha-CLIP pays more attention to user-defined highlighted areas at feature grid space. We check the attention map of \texttt{[CLS]} token in the last transformer block in the vision encoder. The model used for visualization is ViT-L/14 with 16 heads self-attention. For a fair comparison, we use the 5$^{th}$ and 16$^{th}$ heads attention maps for visualization, as we find that these two feature maps are most distinguishable among 16 heads. Results are shown in \cref{fig:more_attn}. This visualization verifies that Alpha-CLIP pays more attention to the area to focus on and more importantly, with no damage to the 2D location information preserved in the feature location of the original CLIP~\cite{CLIP}.

\newpage

\clearpage
\newpage
{
    \small
    \bibliographystyle{ieeenat_fullname}
    \bibliography{main}
}

\end{document}